\pdfoutput=1

\documentclass[11pt]{article}

\usepackage[]{EMNLP2024}

\usepackage{times}
\usepackage{latexsym}

\usepackage[T1]{fontenc}

\usepackage[utf8]{inputenc}

\usepackage{microtype}

\usepackage{inconsolata}

\usepackage{graphicx}
\usepackage{amsmath}
\usepackage{multirow}
\usepackage{subcaption}

\usepackage{booktabs}
\usepackage{xspace}
\usepackage{makecell}

\usepackage{siunitx} %

\usepackage[capitalize,noabbrev]{cleveref}

\usepackage{amssymb}

\usepackage{anyfontsize}

\def\tok100{\mathcal{T}_{100}}

\newlength{\widebarargwidth}
\newlength{\widebarargheight}
\newlength{\widebarargdepth}

\usepackage{fontawesome}

\newcommand{\1}{\mathbb{I}} %

\usepackage{amsmath,amsfonts,bm}

\def\eqref#1{equation~\ref{#1}}

\def\1{\bm{1}}

\DeclareMathAlphabet{\mathsfit}{\encodingdefault}{\sfdefault}{m}{sl}
\SetMathAlphabet{\mathsfit}{bold}{\encodingdefault}{\sfdefault}{bx}{n}

\usepackage{xcolor}

\usepackage{fdsymbol}

\newcommand\blfootnote[1]{%
  \begingroup
\renewcommand\thefootnote{}\footnote{#1}%
  \addtocounter{footnote}{-1}%
  \endgroup
}

\title{Scaling Parameter-Constrained Language Models with Quality Data}

\makeatletter
\def\@fnsymbol#1{\ensuremath{\ifcase#1\or \dagger\or \ddagger\or
\mathsection\or \mathparagraph\or \|\or **\or \dagger\dagger
\or \ddagger\ddagger \else\@ctrerr\fi}}
\makeatother

\author{
Ernie Chang$^{*\spadesuit}$ \,
Matteo Paltenghi$^{*\spadesuit\dagger}$ \,
Yang Li$^{\clubsuit}$ \,
Pin-Jie Lin$^{\vardiamondsuit}$ \,
Changsheng Zhao$^{\spadesuit}$ \,\\
{\bf Patrick Huber$^{\spadesuit}$} \,
{\bf Zechun Liu$^{\spadesuit}$} \,
{\bf Rastislav Rabatin$^{\spadesuit}$} \,
{\bf Yangyang Shi$^{\spadesuit}$} \,
{\bf Vikas Chandra$^{\spadesuit}$} \\
$^{\spadesuit}$AI at Meta \\
$^{\clubsuit}$Iowa State University \\
$^{\vardiamondsuit}$Virginia Tech \\
\texttt{\{erniecyc, mattepalte\}@meta.com, yangli1@iastate.edu, pinjie@vt.edu}
}

\begin{document}

\maketitle

\begin{abstract}

Scaling laws in language modeling traditionally quantify training loss as a function of dataset size and model parameters, providing compute-optimal estimates but often neglecting the impact of data quality on model generalization.
In this paper, we extend the conventional understanding of scaling law by offering a microscopic view of data quality within the original formulation -- \emph{effective training tokens} -- which we posit to be a critical determinant of performance for parameter-constrained language models.
Specifically, we formulate the proposed term of effective training tokens to be a combination of two readily-computed indicators of text:
(i) text diversity and (ii) syntheticity as measured by a teacher model.
We pretrained over $200$ models of 25M to 1.5B parameters on a diverse set of sampled, synthetic data, and estimated the constants that relate text quality, model size, training tokens, and eight reasoning task accuracy scores.
We demonstrated the estimated constants yield +0.83 Pearson correlation with true accuracies, and analyzed it in scenarios involving widely-used data techniques such as data sampling and synthesis which aim to improve data quality.
\blfootnote{$^\ast$ Equal contribution.}
\blfootnote{$^\dagger$ Work done during an internship at Meta.}

\end{abstract}

\section{Introduction}\label{section:introduction}

Recent advancements in language model (LM) development have been significantly influenced by the exploration of scaling laws, which articulate the relationship between training loss, dataset size, and the number of model parameters~\cite{hestness2017deeplearningscalingpredictable,kaplan2020scaling,10.5555/3618408.3618421}. 
These scaling laws have been instrumental in predicting the computational resources necessary for training increasingly large models and have provided a framework for understanding how model performance scales with data and parameters~\cite{hoffmann2022chinchilla,kaplan2020scaling}. 
However, these laws primarily focus on the quantity of data and model size, often underestimating the critical role of data quality in model generalization.

In this work, we challenge the prevailing focus\footnote{\small Both \citet{kaplan2020scaling} and \citet{hoffmann2022chinchilla} formulate scaling law as minimizing loss w.r.t. compute that is parameterized by number of model parameters and training tokens.} on merely increasing data volume and model size by emphasizing the importance of data quality, particularly in scenarios constrained by the number of model parameters. 
We argue that for sub-billion parameter models, the quality of data—or what we term as \emph{effective training tokens} -- plays a more decisive role in model performance than previously recognized. 
This perspective shifts the paradigm from a quantity-centric view to a quality-centric approach in the development of language models.

Further, we provide qualitative measures of standard data refinement techniques including data sampling~\cite{penedo2023refinedweb,wang2024survey,albalak2024survey} and text synthesis~\cite{liu2024best}, applied to a pretraining corpus such as RefinedWeb~\cite{penedo2023refinedweb}. 
This helps to formulate the relationship between the diversity and syntheticity of pretraining data in order to compute the number of \emph{effective training tokens}, which evaluate the impact of data quality in terms of model size and the token number.
Further, we conduct extensive experiments across eight different benchmarks to evaluate the impact of data refinement techniques which allow us to significantly outperform models trained on randomly selected data samples, across a spectrum of model sizes ranging from 25 million to 1.5 billion parameters.

By integrating the notion of \emph{effective token size} into the scaling law formulation, we extend the existing scaling law formulation to better capture the nuances of data quality. 
Our results underscore the pivotal role of high-quality data in training efficient and powerful language models, particularly in parameter-constrained settings.
The contributions of this paper are as follows:
\begin{enumerate}
    \item We extend the conventional scaling law, traditionally expressing training loss as a function of data quantity and model parameters, and incorporate the concept of \emph{effective token size}. 
    This modification emphasizes the importance of data quality in the scaling equation, addressing a critical oversight in previous formulations. 
    \item We investigate the revised scaling law in the context of data refinement techniques such as data selection (e.g. deduplication) and synthesis and investigate their relations to data quality metrics such as diversity and syntheticity. 
    Our finding underscores the potential of data quality, rather than sheer quantity, to enhance model performance.
\end{enumerate}

\section{Background}\label{section:relworks}

Chinchilla scaling law~\cite{hoffmann2022chinchilla} provides a predictive framework for estimating model training loss, considering the number of training tokens and model parameters. Initially designed to identify optimal compute settings for extensive pretraining—a costly and time-consuming endeavor—these laws are crucial for optimizing computational resources. 
Recent studies by \citet{abbas2023semdedupdataefficientlearningwebscale,liu2024best,goyal2024scaling} emphasize the pivotal role of data quality in model pretraining, underscoring the need for revising scaling law formulations.

On the other hand, data refinement can be categorized into \emph{non-transformative} and \emph{transformative} types~\cite{zhao2023survey}. 
Non-transformative refinements involve selective curation of data samples without altering their core characteristics. 
In contrast, transformative refinements generate new text data, rearranging and introducing new tokens, thus impacting training token distributions and data quality. 
This significantly affects the effective number of training tokens used in model training.

In non-transformative refinements, data deduplication is essential for preventing model generalization issues by removing duplicate documents~\cite{lee2022deduplicating,penedo2023refinedweb,tirumala2024d4}. This process not only reduces the number of training tokens but also enhances the quality and effectiveness of the remaining tokens, improving model performance~\cite{muennighoff2024scaling,lee2022deduplicating}.
Data selection, another non-transformative method, involves choosing an optimal data subset from a larger corpus for model training. 
Both approaches aim to enhance model performance, reduce computational costs, and maintain evaluation metric integrity \cite{doi:10.1080/00401706.1975.10489266,10.5555/2380985}.

Transformative refinements, such as synthetic data generation through instructional prompts, are becoming popular~\cite{long2024llms,chung2023increasing,ding2024semcoder}. This approach creates new data to fill existing dataset gaps or introduce new learning scenarios. Integrating synthetic data into large-scale pretraining has significantly improved model robustness and generalization~\cite{li2023textbooksneediiphi15,maini2024rephrasingwebrecipecompute,liu2024best}. Synthetic data generation allows for controlled training dataset expansion, ensuring exposure to diverse inputs and scenarios~\cite{adler2024nemotron}.

Generally, data refinements are crucial in shaping the training landscapes of modern machine learning models, directly influencing training token distribution and quality, thereby enhancing training efficiency and effectiveness in line with scaling laws~\cite{adler2024nemotron}.

\section{Formulating Data Quality}\label{section:quality}

Here we adopt two popular metrics to measuring text quality that are easy to compute on large-scale pretraining data, which is an important considerations when measuring data quality of pretraining sets.

\textbf{Diversity:} 
Following~\citet{shaib2024standardizing}, we utilize the compression ratio, which has been demonstrated to be effective for large-scale pretraining datasets and correlates well with other diversity metrics (Figure~\ref{fig:correlation}). 
Past metrics generally quantify the number of repeated substrings across outputs.
Among these, the token-type ratio is calculated by dividing the count of unique tokens by the total number of tokens in a text. 
To capture the lexical dynamics across varying text lengths, the moving average token type ratios (MATTRs) were introduced, providing a robust measure that is insensitive to text length \cite{Covington2010CuttingTG}. 
This metric focuses on the frequency of individual word repetition within text segments and does not account for longer repeated sequences.

\begin{figure}[t]
\centering    
\includegraphics[width=0.75\columnwidth]{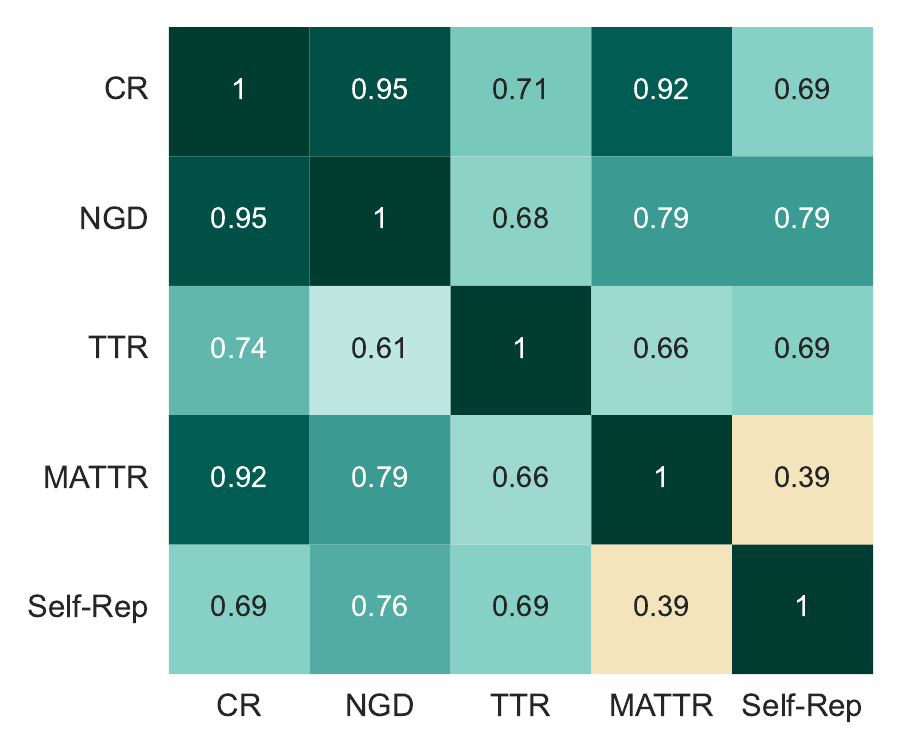}
    \caption{\small 
Correlations between text diversity scores on 1\% of RefinedWeb~\cite{penedo2023refinedweb}. Similar to~\cite{shaib2024standardizing}, compression ratio (CR) correlates strongly with most other diversity metrics. }
\label{fig:correlation}
\end{figure}

To address longer sequences, the concept of token-type ratio has been expanded through the introduction of \emph{n-gram diversity}, as explored in recent studies \cite{padmakumar-etal-2023-investigating,meister-etal-2023-locally,li-etal-2016-diversity}. 
Additionally, the metric of self-repetition has been developed to assess the tendency of language models to repeat long n-grams across different outputs \cite{salkar2022self}, which measures language model's inclination towards redundancy in longer sequences.
To this end, we employ text compression algorithms designed to identify redundancy in sequences of variable length. 
We use gzip~\cite{gailly1992gnu} to compress the concatenated text of all outputs generated by a model. 
The compression ratio, which compares the size of the original file to that of the compressed file, serves as an indicator of redundancy:
\begin{equation}
CR(D) = \frac{\text{Original size of } D\oplus \text{ (in bytes)}}{\text{Compressed size of } D\oplus \text{ (in bytes)}} \nonumber 
\end{equation}
\begin{equation}
\label{eq:diversity}
\text{Dr}(D) = \text{CR}^{-1}(D)
\end{equation}
High compression ratios suggest greater redundancy, indicating lower diversity within the text data.
Therefore, diversity is defined as $\text{Dr}(D)$, where higher means more diverse text.

\textbf{Syntheticity:} We estimate the syntheticity of data points in our dataset using the perplexity metric, which is calculated with a teacher-model, i.e. Llama-2 7B chat~\cite{touvron2023llama}\footnote{This smaller pretrained model is selected due to practical concerns over the total scoring time.}.
This model choice is strategic because teacher models are known for their robust performance across a variety of benchmarks and their alignment with safety choices, making them reliable for general evaluations without needing to tailor them to specific downstream tasks.
Perplexity, in this context, measures how well the teacher model predicts a sequence of subword tokens, with lower values indicating higher predictability and, by extension, higher syntheticity.
A low perplexity score suggests that the data point is well-represented by the model's learned patterns, which could indirectly indicate that it is more relevant or useful for similar tasks or applications.
Hence syntheticity is inversely proportional to perplexity and is then defined as follows:
\begin{align} \label{eq:ppl}
\text{S}(D) = \exp^{-1}\left(-\frac{1}{M} \sum_{i=1}^{M} \log P(w_i | w_{<i})\right)
\end{align}
The formula above calculates the inverse of the exponential of the negative average log-likelihood of predicting each subword token in the document $D$, given all previous tokens.
This quantifies how expected the tokens are, given the model's current knowledge state, thus providing a direct measure of how typical or atypical the sequence is within the context of the teacher model.

\section{Scaling Law with Data Quality}\label{sec:scaling_law_with_data_quality}
We propose to modify the third approach of the Chinchilla scaling law~\cite{hoffmann2022chinchilla} which originally models the losses in training large language models with the functional form $E + \frac{A}{N^\alpha} + \frac{B}{D^\beta}$ with the constants: ($E = 1.89, A = 463.3, \alpha = 0.345, B = 12530, \beta = 0.452$)\footnote{\small Later work from~\citet{besiroglu2024chinchilla} re-estimated the constants from the original Chinchilla scaling law with more plausible confidence level.}. 
In this formulation, ($E$) represents the baseline loss, akin to the entropy of natural text under an ideal generative process, setting the theoretical minimum loss achievable with data $D$ and model parameter $N$.

In this work, we model the zero-shot accuracy on common sense reasoning as we postulate that the score provides an indication on how much reasoning ability a given data $D$ could possibly instill. 
To incorporate data quality into this framework, we propose to use a quality term $Q$ to provide a quality-adjusted number of training tokens ($D_q$), combining Eq.~\ref{eq:diversity} and Eq.~\ref{eq:ppl}:
\begin{align} 
D_{\text{q}} &= D \cdot \exp(c_1 \cdot \text{diversity} + c_2 \cdot \text{syntheticity}) \nonumber \\
&= D \cdot \underbrace{\exp(c_1 \cdot \text{Dr}(D) + c_2 \cdot \text{S}(D)}_{\text{Scaling factor Q} }) \label{eq:scaling_factor}
\end{align}
where ($c_1$) and ($c_2$) are scaling factors that adjust ($D_q$) to account for the syntheticity and diversity of the training tokens.
Here we revise the scaling law to predict the average zero-shot accuracy $G$ across eight reasoning tasks\footnote{\small We employ ARC-easy, ARC-challenge~\cite{clark2018arc}, BoolQ~\cite{clark2019boolq}, PIQA~\cite{bisk2020piqa}, SIQA~\cite{sap2019siqa}, HellaSwag~\cite{zellers2019hellaswag}, OBQA~\cite{mihaylov2018obqa}, and WinoGrande~\cite{sakaguchi2021winogrande} as the tasks that define the score $\hat{G}(N,D)$.} instead of loss as given by:
\begin{equation}\label{eq:scaling_law}
\begin{split}
\hat{G}(N,D) &= \mathcal{R}\left(E + \frac{A}{N^\alpha} + \frac{B}{D_{\text{q}}^\beta}\right) \\
\mathcal{R}(x) &= \min(\max(x, 0), 1)
\end{split}
\end{equation}

This revision integrates the quality-adjusted number of training tokens ($D_{\text{q}}$)
into the accuracy function, allowing for a more nuanced understanding of how data quality impacts model training and performance.

\section{Data Refinement: A Case Study}\label{section:case_study_refinement}

We explore two prevalent data refinement techniques aimed at enhancing data quality: data selection and data synthesis. These methods have become standard practices in the preparation of pretraining datasets, significantly influencing text diversity and syntheticity and downstream performance as shown in various studies~\cite{abdin2024phi,albalak2024survey}.

\begin{figure}[ht]
\centering    
\includegraphics[width=1.1\columnwidth]{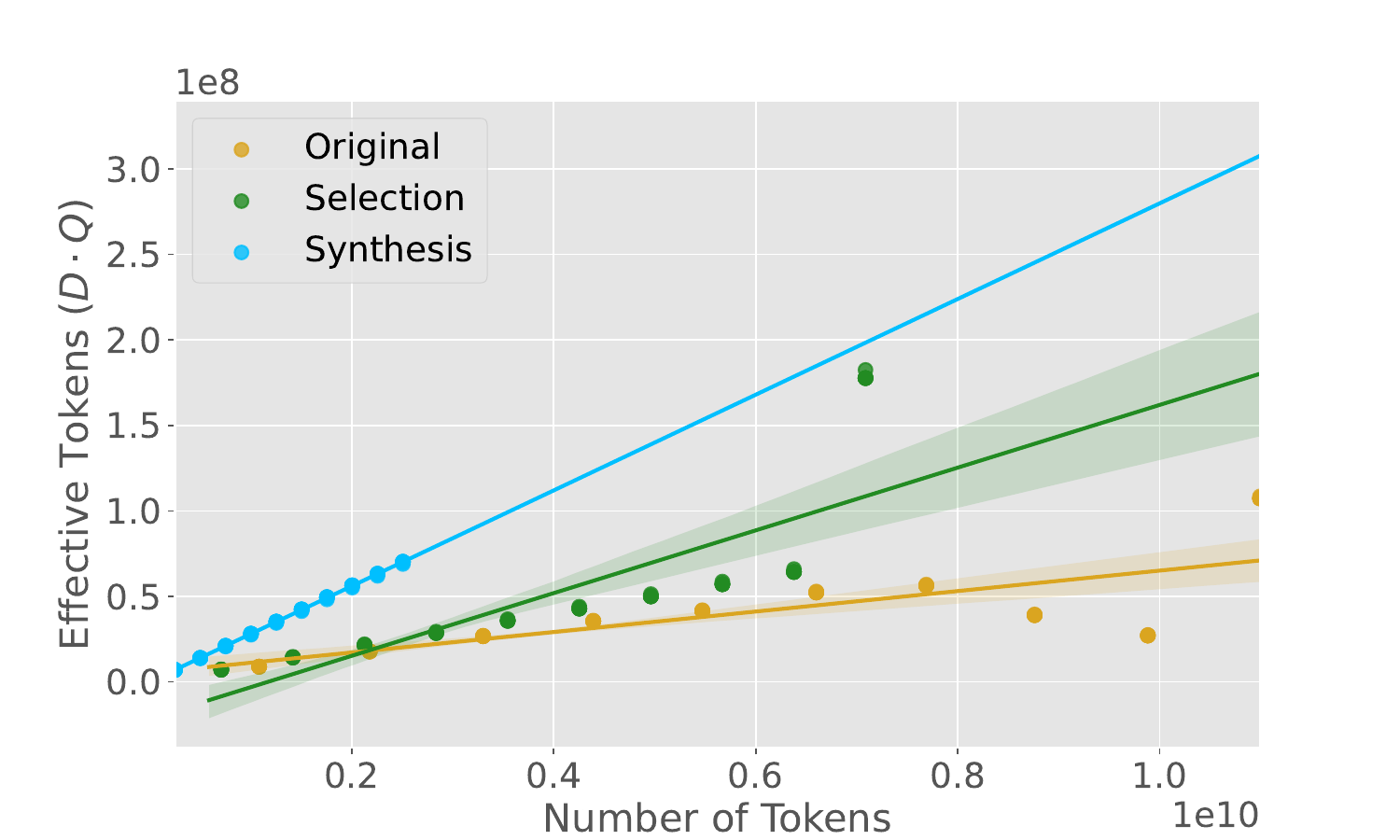}
    \caption{\small This plot illustrates the impact of various data refinement techniques on the effective token count ($D_q$) as the number of tokens is scaled up. 
    Experiments were performed with RefinedWeb~\cite{penedo2023refinedweb} data.
}
\label{fig:domain}
\end{figure}

To put them in context, we present a comparative analysis in Figure~\ref{fig:domain}, which displays the relationship between effective token counts $D_q$ and the total number of tokens $D$. 
It clearly demonstrates that data synthesis has a more substantial impact on increasing the effective token count compared to data selection and the use of original datasets. 
This underscores the value of synthesis in optimizing data quality for model training.

\subsection{Data Selection}

\paragraph{Coreset Selection.}
One way to create a higher quality dataset is via importance sampling~\cite{xie2023data,wang2018dataset}, which transformed input data into n-gram based feature vectors and compares the feature distributions between the raw and target datasets and assigning importance weights to each example.

\begin{figure*}[t]
    \centering

\includegraphics[width=\textwidth]{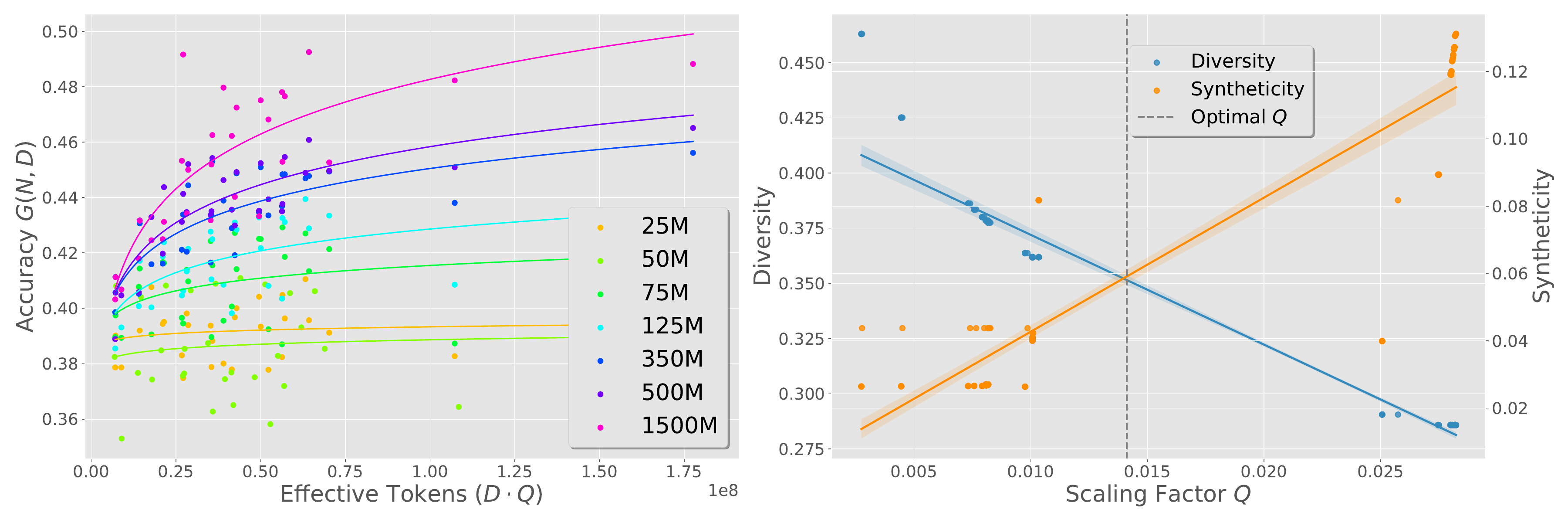}
    \caption{\small
    Plots of revised scaling law with qualitative data measurements. 
    Left: Plot of averaged accuracy against effective tokens $D_q$ where $
D_q = D \cdot \exp(c_1 \cdot \text{Dr}(D) + c_2 \cdot \text{S}(D))$. The accuracy values are the reference values.
    Right: Impact of scaling factor $Q$ on both diversity and syntheticity. 
    Interestingly, we found that diversity needs to be reduced while syntheticity needs to be increased for scaling factor to go up, which can then improve overall accuracy.  
We include the constant values in Table~\ref{tab:param-estimates}.}
    \label{fig:learning_curve}
\end{figure*}

This selectively enhance the dataset's syntheticity and directly influenced the \(D_q\) term in the revised scaling law, increasing the syntheticity factor without compromising on diversity. 
While this approach assumes the knowledge of target applications, but it also allows us to easily explore the impact of having more in-domain data on the data quality and losses.

\paragraph{Text Deduplication.}
An orthogonal approach is text deduplication~\cite{sorscher2022beyond,penedo2023refinedweb,penedo2024fineweb} which removes redundant data, ensuring a balanced dataset that does not favor frequently occurring examples. 
This method modulates the diversity and quality of the dataset, which is crucial for robust model training. 
The deduplication process effectively controlled the \(D_q\) term by filtering out excessive redundancy, which could lead to overfitting if left unchecked.

\subsection{Synthetic Data}
In \emph{transformative} data refinement, one popular approach is to utilize a teacher model trained on a diverse and comprehensive dataset to generate synthetic data~\cite{narayan2024cookbook,abdin2024phi}.
We provided the instruction prompts in the appendix, which aim to paraphrased pretraining documents. 
In general, the synthetic data broadened the diversity of the dataset and introduced more complex token patterns, which can lead to improved model performance, particularly in providing complex scenarios that were not well-represented in the original dataset.

\section{Experimental Setup}
\textbf{Network and Training Details.} For all experiments, we pretrain the decoder-only transformer using causal language modeling objectives on selected datasets, where model weights were randomly initialized.
We evaluated with the language models of sizes $\{25, 50, 75, 125, 350, 500\}$M and 1.5B parameters which allowed us to explore how model capacity impacts the final results.
Pretraining was conducted on a distributed computing setup with 32 GPUs across 4 nodes, each equipped with an H100 graphics card.

\paragraph{Data Preparations.}

For our evaluations, we benchmarked the models across eight common sense reasoning tasks in a zero-shot setting, including ARC-easy, ARC-challenge~\cite{clark2018arc}, BoolQ~\cite{clark2019boolq}, PIQA~\cite{bisk2020piqa}, SIQA~\cite{sap2019siqa}, HellaSwag~\cite{zellers2019hellaswag}, OBQA~\cite{mihaylov2018obqa}, and WinoGrande~\cite{sakaguchi2021winogrande}. We selected a random sample of 16M JSON objects from RefinedWeb, formatted in JSONL. The dataset was then segmented into increments of 10\% ranging from 10\% to 100\% of the data, and used to pre-train seven different model sizes.

The token counts for these models were set at $\{2,4,6,8,10\}$ billion tokens, with each model trained using an equivalent amount of computational resources. Our hardware setup included 4 nodes, each equipped with 8 GPUs, running for 100,000 steps with a context length of 2048 and a batch size of 16. This configuration ensured that each model was sufficiently trained, with the largest dataset undergoing approximately 9.5 epochs and the smallest dataset experiencing about 48.1 epochs. Intermediate model sizes were trained for epochs falling between these two extremes.

To ensure a diverse range of training data, we constructed several datasets from multiple sources, including \emph{random} data (8B tokens), \emph{selected} data (7B tokens), and \emph{synthetic} data (2B tokens). The \emph{selected} data was curated based on the evaluation set of the eight tasks using importance sampling~\cite{xie2023data}, while the \emph{synthetic} data was generated through instructional prompts aimed at paraphrasing each pretraining document. In contrast, the \emph{random} data was noted for its high diversity but low syntheticity, as discussed in \cref{section:quality}. Conversely, the \emph{synthetic} data exhibited the lowest diversity but the highest syntheticity score.

\begin{table}[th]
\centering
\label{tab:param-estimates}
\small
\resizebox{0.9\columnwidth}{!}{
\begin{tabular}{@{}lcc@{}}
\toprule
Parameter & \citet{besiroglu2024chinchilla} & Ours \\
\midrule
$A$ & 482.01 (\textit{124.58}) & -0.8546  \\
$B$ & 2085.43 (\textit{1293.23}) & -18.3078  \\
$E$ & 1.8172 (\textit{0.03}) & 1.1400  \\
$\alpha$ & 0.3478 (\textit{0.02}) & 0.0450  \\
$\beta$ & 0.3658 (\textit{0.02}) & 0.3683  \\
$c_1$ & - & -12.7756  \\
$c_2$ & - & 0.6369  \\
Data points & 240 & 210 \\
\bottomrule
\end{tabular}}
\caption{\small Parameter estimates and their standard errors. The standard errors are shown in parentheses and are obtained by bootstrapping. We show the estimates from \citet{besiroglu2024chinchilla} (re-estimated from \citet{hoffmann2022chinchilla}) for comparison and added the constants $c_1$ and $c_2$ for text diversity and syntheticity respectively.}
\label{tab:param-estimates}
\end{table}

\section{Discussions}

By over 200 training runs, we re-estimate all the constants which we show in Table~\ref{tab:param-estimates}. 
Here we first discuss the estimation of constants that relate to accuracy and the rest of the scaling parameters in Eq.~\ref{eq:scaling_law}. 
In particular, we discuss the scaling factor $Q$ and how it can be applied to pretraining scenarios.

\paragraph{Correlation Strength of Estimated Constants.}

\begin{figure}[t]
\centering    
\includegraphics[width=1\columnwidth]{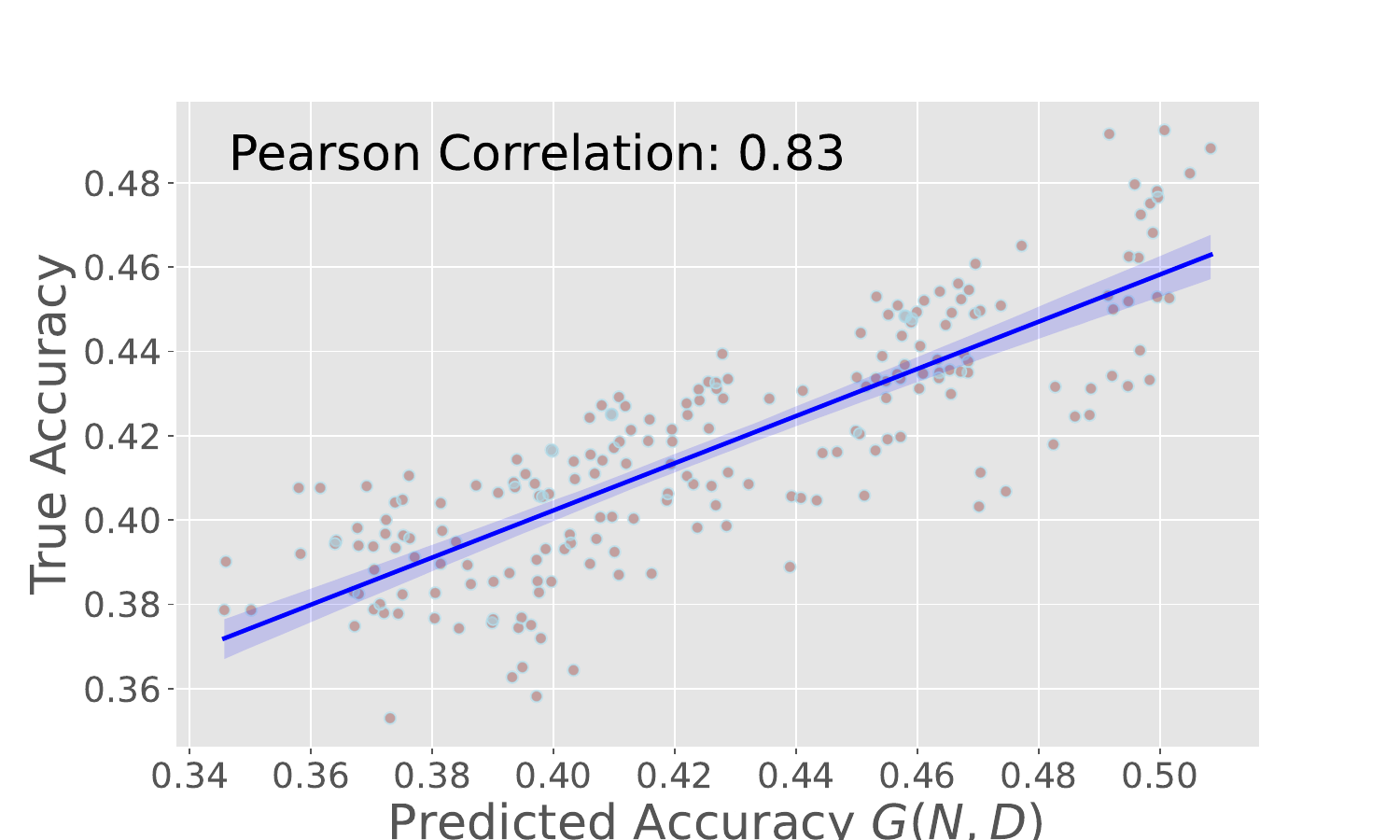}
    \caption{\small This plot illustrates correlation between the predicted accuracy $G(N, D)$ and the true accuracy of RefinedWeb data. The Pearson correlation is +0.83. 
}
\label{fig:pearson}
\end{figure}

In Table~\ref{tab:param-estimates}, we show the estimated constants for the scaling law Eq.\ref{eq:scaling_law} and the proposed scaling factor term Eq.\ref{eq:scaling_factor}. 
The constants were estimated with the nonlinear least-squares method with the Scipy optimizer\footnote{\url{https://docs.scipy.org/doc/scipy/reference/generated/scipy.optimize.curve_fit.html}}, where the initial guesses were the original Chinchilla scaling law constants in~\citet{hoffmann2022chinchilla}, and the maximum number of function calls was set as $2000$. 
To validate our estimated constants, we provide a predicted vs. true accuracy plot and the Pearson correlation in Figure~\ref{fig:pearson}. 
This gives us ideas on how strongly these constants are correlated to the training set used to estimate our revised scaling formulation.
Strikingly, this amounts to the correlation strength of +$0.83$ across all model sizes and data samples.
We attribute the robustness of the formulation to the use of data-agnostic compression ratio and a reasonably-capable language model as teacher. 
\paragraph{How to Improve Data Quality for Better Models?}
\begin{figure}[t]
\centering    
\includegraphics[width=0.9\columnwidth]{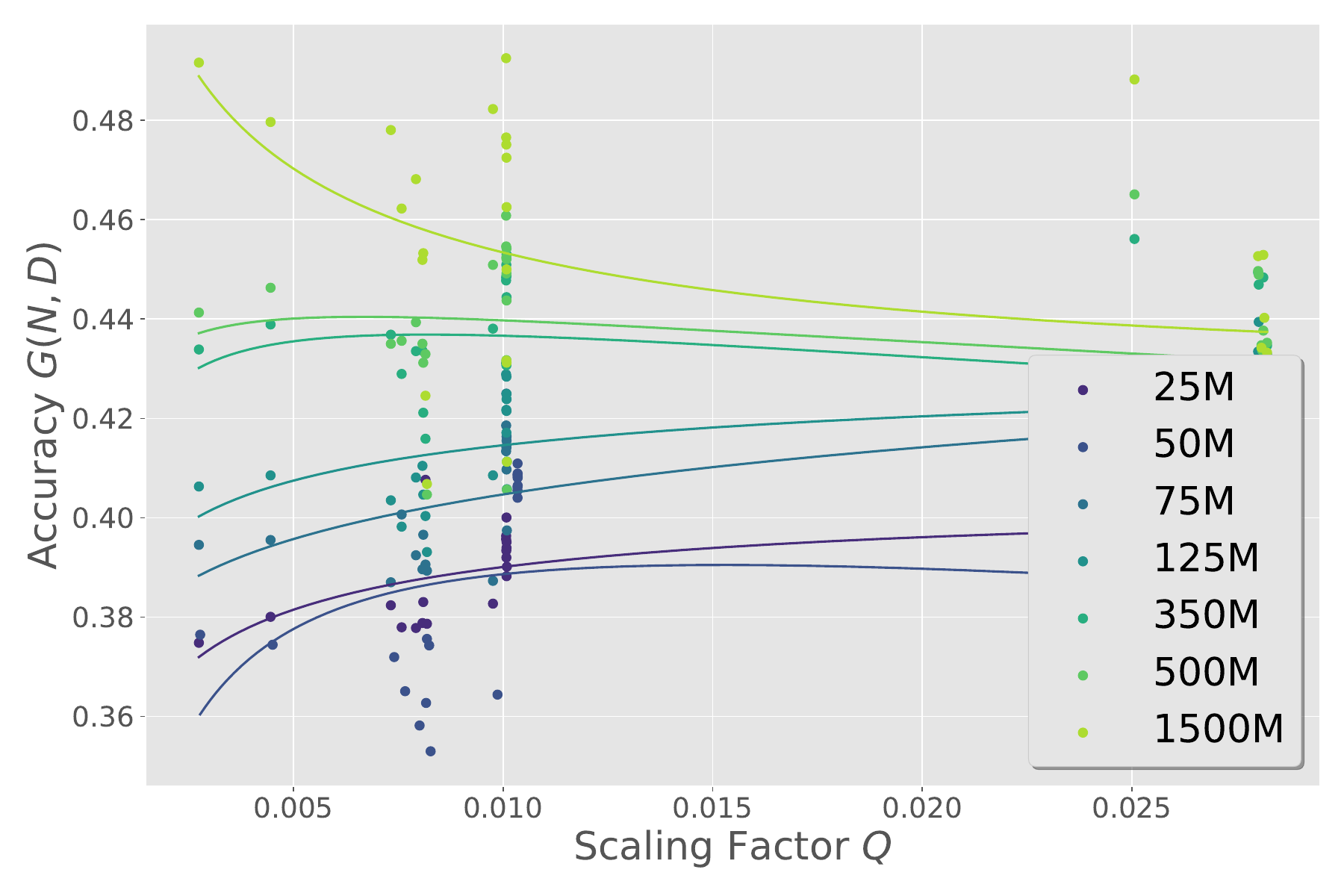}
    \caption{\small This plot illustrates correlation between the accuracy and the scaling factor $Q$ across all model sizes, which shows that scaling up the value of $Q$ improves accuracy up to a point, where then the token number becomes dominant.
}
\label{fig:correlation}
\end{figure}

In the left plot of Figure~\ref{fig:learning_curve}, we first explore the impact of effective tokens on model accuracy. It is evident that an increase in effective tokens correlates with higher accuracy. However, the influence of the scaling factor $Q$ varies across different models. Notably, the impact of data quality is more pronounced in smaller model sizes ranging from 25M to 500M, and it gradually levels off as the value of scaling factor $Q$ increases, eventually reaching a point where effective tokens $D_q$ are predominantly determined by the sheer number of tokens.
Additionally, we examine the interplay between the scaling factor $Q$, diversity, and syntheticity in the right plot of Figure~\ref{fig:learning_curve}. Several key observations emerge:
\begin{enumerate}
    \item There is an inverse relationship between diversity and syntheticity, which is expected as synthetic data generated by language models tends to be less diverse.
    \item Less diverse data increases the value of the scaling factor; conversely, more synthetic data tends to elevate scaling factor $Q$.
    \item However, when the curves of diversity and syntheticity converge, the influence of the scaling factor $Q$ on accuracy improvement becomes negligible.
\end{enumerate}

\paragraph{Data Quality Scaling is Token Quantity Bound.}

These insights establish some basic guidelines:
To enhance data quality in smaller models, introducing synthetic data can be beneficial, as it typically yields less diverse but more synthetic data with a higher scaling factor $Q$. 
However, it is crucial for training practitioners to recognize that while increasing text syntheticity can scale up data quality, the total token count ultimately plays a more dominant role in improving model accuracy in larger models that are more data-hungry (e.g. greater than 1.5B in our experiments), as illustrated in Figure~\ref{fig:correlation}.

\section{Conclusion and Future Works}\label{section:conclusion}
In this paper, we revisited traditional scaling laws in language modeling that often overlook the critical impact of data quality on model generalization. 
We introduced the concept of effective training tokens, emphasizing its significance in enhancing model performance, particularly for models with constrained parameters, in order to offer a more precise understanding of data quality's role in model scaling. 
Our findings highlight the pivotal role of data quality and pave the way for developing more efficient and compact language models suitable for on-device applications.

\section*{Limitations}

While our revised scaling law incorporating effective training tokens offers a nuanced understanding of data quality, a significant limitation arises from the number of sample points required to accurately estimate the constants within the law. 
The precision of these constants is crucial as they directly influence the model's performance predictions and generalizations. 
However, obtaining a sufficient number of diverse and representative sample points to robustly estimate these constants is challenging. 
This limitation is particularly pronounced in scenarios involving rare or complex data characteristics, where the availability of adequate and varied training examples is limited. Consequently, the reliability of our scaling law under these conditions may be compromised, necessitating further research and potentially more sophisticated sampling techniques to enhance the robustness of our estimates.

\section*{Ethics Statement}

In this study, we explore the impact of data quality on language model performance by introducing the concept of \emph{effective training tokens}. 
Our experiments, conducted on a diverse set of sampled and synthetic data, adhere to rigorous standards to ensure the reproducibility and reliability of our findings. While our research utilizes datasets that are well-established within the academic community, the application of our findings to sensitive or private datasets must be approached with strict ethical considerations and robust privacy safeguards. 
Additionally, the methodologies proposed for enhancing data quality, such as text diversity and fidelity assessments, should be applied judiciously to avoid unintended biases or ethical dilemmas. 
As we push the boundaries of model efficiency and performance, it is imperative to balance these advancements with careful consideration of their broader implications, including the potential increase in computational demands and its associated environmental impact.

\bibliography{anthology,bibtex/coreset,bibtex/custom,bibtex/datasets,bibtex/nas-overview,bibtex/pretrained,bibtex/zero_nas,bibtex/mllm,bibtex/sampling}

\begin{thebibliography}{28}
\expandafter\ifx\csname natexlab\endcsname\relax\def\natexlab#1{#1}\fi

\bibitem[{Brown et~al.(2020)Brown, Mann, Ryder, Subbiah, Kaplan, Dhariwal,
  Neelakantan, Shyam, Sastry, Askell et~al.}]{GPT3}
Tom Brown, Benjamin Mann, Nick Ryder, Melanie Subbiah, Jared~D Kaplan, Prafulla
  Dhariwal, Arvind Neelakantan, Pranav Shyam, Girish Sastry, Amanda Askell,
  et~al. 2020.
\newblock Language models are few-shot learners.
\newblock \emph{Advances in neural information processing systems},
  33:1877--1901.

\bibitem[{Chalkidis et~al.(2022)Chalkidis, Jana, Hartung, Bommarito,
  Androutsopoulos, Katz, and Aletras}]{LexGLUE}
Ilias Chalkidis, Abhik Jana, Dirk Hartung, Michael Bommarito, Ion
  Androutsopoulos, Daniel Katz, and Nikolaos Aletras. 2022.
\newblock \href {https://doi.org/10.18653/v1/2022.acl-long.297} {{L}ex{GLUE}: A
  benchmark dataset for legal language understanding in {E}nglish}.
\newblock In \emph{Proceedings of the 60th Annual Meeting of the Association
  for Computational Linguistics (Volume 1: Long Papers)}, pages 4310--4330,
  Dublin, Ireland. Association for Computational Linguistics.

\bibitem[{Dao et~al.(2022)Dao, Fu, Ermon, Rudra, and
  R{\'e}}]{dao2022flashattention}
Tri Dao, Dan Fu, Stefano Ermon, Atri Rudra, and Christopher R{\'e}. 2022.
\newblock Flashattention: Fast and memory-efficient exact attention with
  io-awareness.
\newblock \emph{Advances in Neural Information Processing Systems},
  35:16344--16359.

\bibitem[{Devlin et~al.(2018)Devlin, Chang, Lee, and Toutanova}]{BERT}
Jacob Devlin, Ming-Wei Chang, Kenton Lee, and Kristina Toutanova. 2018.
\newblock {BERT}: Pre-training of deep bidirectional transformers for language
  understanding.
\newblock \emph{arXiv preprint arXiv:1810.04805}.

\bibitem[{Gee et~al.(2022)Gee, Zugarini, Rigutini, and Torroni}]{FVT}
Leonidas Gee, Andrea Zugarini, Leonardo Rigutini, and Paolo Torroni. 2022.
\newblock \href {https://aclanthology.org/2022.emnlp-industry.41} {Fast
  vocabulary transfer for language model compression}.
\newblock In \emph{Proceedings of the 2022 Conference on Empirical Methods in
  Natural Language Processing: Industry Track}, pages 409--416, Abu Dhabi, UAE.
  Association for Computational Linguistics.

\bibitem[{Gupta et~al.(2015)Gupta, Agrawal, Gopalakrishnan, and
  Narayanan}]{Precision}
Suyog Gupta, Ankur Agrawal, Kailash Gopalakrishnan, and Pritish Narayanan.
  2015.
\newblock Deep learning with limited numerical precision.
\newblock In \emph{International conference on machine learning}, pages
  1737--1746. PMLR.

\bibitem[{Gurulingappa et~al.(2012)Gurulingappa, Rajput, Roberts, Fluck,
  Hofmann-Apitius, and Toldo}]{ADE}
Harsha Gurulingappa, Abdul~Mateen Rajput, Angus Roberts, Juliane Fluck, Martin
  Hofmann-Apitius, and Luca Toldo. 2012.
\newblock \href {https://doi.org/https://doi.org/10.1016/j.jbi.2012.04.008}
  {Development of a benchmark corpus to support the automatic extraction of
  drug-related adverse effects from medical case reports}.
\newblock \emph{Journal of Biomedical Informatics}, 45(5):885 -- 892.
\newblock Text Mining and Natural Language Processing in Pharmacogenomics.

\bibitem[{Hinton et~al.(2015)Hinton, Vinyals, Dean et~al.}]{Distillation}
Geoffrey Hinton, Oriol Vinyals, Jeff Dean, et~al. 2015.
\newblock Distilling the knowledge in a neural network.
\newblock \emph{arXiv preprint arXiv:1503.02531}, 2(7).

\bibitem[{Jiao et~al.(2020)Jiao, Yin, Shang, Jiang, Chen, Li, Wang, and
  Liu}]{TinyBERT}
Xiaoqi Jiao, Yichun Yin, Lifeng Shang, Xin Jiang, Xiao Chen, Linlin Li, Fang
  Wang, and Qun Liu. 2020.
\newblock {TinyBERT}: Distilling {BERT} for natural language understanding.
\newblock In \emph{Findings of the Association for Computational Linguistics:
  EMNLP 2020}, pages 4163--4174.

\bibitem[{Kudo and Richardson(2018)}]{SentencePiece}
Taku Kudo and John Richardson. 2018.
\newblock Sentencepiece: A simple and language independent subword tokenizer
  and detokenizer for neural text processing.
\newblock \emph{arXiv preprint arXiv:1808.06226}.

\bibitem[{Kumar and Thawani(2022)}]{MBPE}
Dipesh Kumar and Avijit Thawani. 2022.
\newblock Bpe beyond word boundary: How not to use multi word expressions in
  neural machine translation.
\newblock In \emph{Proceedings of the Third Workshop on Insights from Negative
  Results in NLP}, pages 172--179.

\bibitem[{Lample et~al.(2018)Lample, Ott, Conneau, Denoyer, and
  Ranzato}]{Phrase}
Guillaume Lample, Myle Ott, Alexis Conneau, Ludovic Denoyer, and Marc'Aurelio
  Ranzato. 2018.
\newblock Phrase-based \& neural unsupervised machine translation.
\newblock \emph{arXiv preprint arXiv:1804.07755}.

\bibitem[{Michel et~al.(2019)Michel, Levy, and Neubig}]{16Heads}
Paul Michel, Omer Levy, and Graham Neubig. 2019.
\newblock Are sixteen heads really better than one?
\newblock \emph{Advances in neural information processing systems}, 32.

\bibitem[{Mu et~al.(2023)Mu, Li, and Goodman}]{mu2023learning}
Jesse Mu, Xiang~Lisa Li, and Noah Goodman. 2023.
\newblock Learning to compress prompts with gist tokens.
\newblock \emph{arXiv preprint arXiv:2304.08467}.

\bibitem[{OpenAI(2023)}]{openai2023gpt4}
OpenAI. 2023.
\newblock \href {http://arxiv.org/abs/2303.08774} {Gpt-4 technical report}.

\bibitem[{Otani et~al.(2020)Otani, Ozaki, Zhao, Li, St~Johns, and Levin}]{MWE}
Naoki Otani, Satoru Ozaki, Xingyuan Zhao, Yucen Li, Micaelah St~Johns, and Lori
  Levin. 2020.
\newblock \href {https://doi.org/10.18653/v1/2020.emnlp-main.360}
  {Pre-tokenization of multi-word expressions in cross-lingual word
  embeddings}.
\newblock In \emph{Proceedings of the 2020 Conference on Empirical Methods in
  Natural Language Processing (EMNLP)}, pages 4451--4464, Online. Association
  for Computational Linguistics.

\bibitem[{Petrov et~al.(2023)Petrov, La~Malfa, Torr, and
  Bibi}]{petrov2023language}
Aleksandar Petrov, Emanuele La~Malfa, Philip~HS Torr, and Adel Bibi. 2023.
\newblock Language model tokenizers introduce unfairness between languages.
\newblock \emph{arXiv preprint arXiv:2305.15425}.

\bibitem[{Sanh et~al.(2019)Sanh, Debut, Chaumond, and Wolf}]{DistilBERT}
Victor Sanh, Lysandre Debut, Julien Chaumond, and Thomas Wolf. 2019.
\newblock \href {https://doi.org/10.48550/ARXIV.1910.01108} {Distilbert, a
  distilled version of bert: smaller, faster, cheaper and lighter}.

\bibitem[{Schuster and Nakajima(2012)}]{WordPiece}
Mike Schuster and Kaisuke Nakajima. 2012.
\newblock Japanese and korean voice search.
\newblock In \emph{2012 IEEE international conference on acoustics, speech and
  signal processing (ICASSP)}, pages 5149--5152. IEEE.

\bibitem[{Sennrich et~al.(2016)Sennrich, Haddow, and Birch}]{BPE}
Rico Sennrich, Barry Haddow, and Alexandra Birch. 2016.
\newblock Neural machine translation of rare words with subword units.
\newblock In \emph{Proceedings of the 54th Annual Meeting of the Association
  for Computational Linguistics (Volume 1: Long Papers)}, pages 1715--1725.

\bibitem[{Sharma et~al.(2019)Sharma, Li, and Wang}]{PATENT}
Eva Sharma, Chen Li, and Lu~Wang. 2019.
\newblock \href {https://doi.org/10.18653/v1/P19-1212} {{BIGPATENT}: A
  large-scale dataset for abstractive and coherent summarization}.
\newblock In \emph{Proceedings of the 57th Annual Meeting of the Association
  for Computational Linguistics}, pages 2204--2213, Florence, Italy.
  Association for Computational Linguistics.

\bibitem[{Shen et~al.(2020)Shen, Dong, Ye, Ma, Yao, Gholami, Mahoney, and
  Keutzer}]{QBERT}
Sheng Shen, Zhen Dong, Jiayu Ye, Linjian Ma, Zhewei Yao, Amir Gholami,
  Michael~W Mahoney, and Kurt Keutzer. 2020.
\newblock {Q-BERT}: Hessian based ultra low precision quantization of {BERT}.
\newblock In \emph{Proceedings of the AAAI Conference on Artificial
  Intelligence}, volume~34, pages 8815--8821.

\bibitem[{Sun et~al.(2020)Sun, Yu, Song, Liu, Yang, and Zhou}]{MobileBERT}
Zhiqing Sun, Hongkun Yu, Xiaodan Song, Renjie Liu, Yiming Yang, and Denny Zhou.
  2020.
\newblock {MobileBERT}: a compact task-agnostic {BERT} for resource-limited
  devices.
\newblock \emph{arXiv preprint arXiv:2004.02984}.

\bibitem[{Touvron et~al.(2023)Touvron, Lavril, Izacard, Martinet, Lachaux,
  Lacroix, Rozi{\`e}re, Goyal, Hambro, Azhar et~al.}]{touvron2023llama}
Hugo Touvron, Thibaut Lavril, Gautier Izacard, Xavier Martinet, Marie-Anne
  Lachaux, Timoth{\'e}e Lacroix, Baptiste Rozi{\`e}re, Naman Goyal, Eric
  Hambro, Faisal Azhar, et~al. 2023.
\newblock Llama: Open and efficient foundation language models.
\newblock \emph{arXiv preprint arXiv:2302.13971}.

\bibitem[{Tuggener et~al.(2020)Tuggener, von D{\"a}niken, Peetz, and
  Cieliebak}]{LEDGAR}
Don Tuggener, Pius von D{\"a}niken, Thomas Peetz, and Mark Cieliebak. 2020.
\newblock Ledgar: A large-scale multi-label corpus for text classification of
  legal provisions in contracts.
\newblock In \emph{Proceedings of the 12th Language Resources and Evaluation
  Conference}, pages 1235--1241.


\bibitem[{Vaswani et~al.(2017)Vaswani, Shazeer, Parmar, Uszkoreit, Jones,
  Gomez, Kaiser, and Polosukhin}]{Transformer}
Ashish Vaswani, Noam Shazeer, Niki Parmar, Jakob Uszkoreit, Llion Jones,
  Aidan~N Gomez, {\L}ukasz Kaiser, and Illia Polosukhin. 2017.
\newblock Attention is all you need.
\newblock \emph{Advances in neural information processing systems}, 30.


\bibitem[{Wang et~al.(2020)Wang, Wei, Dong, Bao, Yang, and Zhou}]{MiniLM}
Wenhui Wang, Furu Wei, Li~Dong, Hangbo Bao, Nan Yang, and Ming Zhou. 2020.
\newblock Minilm: Deep self-attention distillation for task-agnostic
  compression of pre-trained transformers.
\newblock \emph{Advances in Neural Information Processing Systems},
  33:5776--5788.

\bibitem[{Zhu and Gupta(2017)}]{Prune}
Michael Zhu and Suyog Gupta. 2017.
\newblock To prune, or not to prune: exploring the efficacy of pruning for
  model compression.
\newblock \emph{arXiv preprint arXiv:1710.01878}.

\end{thebibliography}


\begin{thebibliography}{44}
\expandafter\ifx\csname natexlab\endcsname\relax\def\natexlab#1{#1}\fi

\bibitem[{Abbas et~al.(2023)Abbas, Tirumala, Simig, Ganguli, and Morcos}]{abbas2023semdedupdataefficientlearningwebscale}
Amro Abbas, Kushal Tirumala, Dániel Simig, Surya Ganguli, and Ari~S. Morcos. 2023.
\newblock \href {http://arxiv.org/abs/2303.09540} {Semdedup: Data-efficient learning at web-scale through semantic deduplication}.

\bibitem[{Abdin et~al.(2024)Abdin, Jacobs, Awan, Aneja, Awadallah, Awadalla, Bach, Bahree, Bakhtiari, Behl et~al.}]{abdin2024phi}
Marah Abdin, Sam~Ade Jacobs, Ammar~Ahmad Awan, Jyoti Aneja, Ahmed Awadallah, Hany Awadalla, Nguyen Bach, Amit Bahree, Arash Bakhtiari, Harkirat Behl, et~al. 2024.
\newblock Phi-3 technical report: A highly capable language model locally on your phone.
\newblock \emph{arXiv preprint arXiv:2404.14219}.

\bibitem[{Adler et~al.(2024)Adler, Agarwal, Aithal, Anh, Bhattacharya, Brundyn, Casper, Catanzaro, Clay, Cohen et~al.}]{adler2024nemotron}
Bo~Adler, Niket Agarwal, Ashwath Aithal, Dong~H Anh, Pallab Bhattacharya, Annika Brundyn, Jared Casper, Bryan Catanzaro, Sharon Clay, Jonathan Cohen, et~al. 2024.
\newblock Nemotron-4 340b technical report.
\newblock \emph{arXiv preprint arXiv:2406.11704}.

\bibitem[{Aghajanyan et~al.(2023)Aghajanyan, Yu, Conneau, Hsu, Hambardzumyan, Zhang, Roller, Goyal, Levy, and Zettlemoyer}]{10.5555/3618408.3618421}
Armen Aghajanyan, Lili Yu, Alexis Conneau, Wei-Ning Hsu, Karen Hambardzumyan, Susan Zhang, Stephen Roller, Naman Goyal, Omer Levy, and Luke Zettlemoyer. 2023.
\newblock Scaling laws for generative mixed-modal language models.
\newblock In \emph{Proceedings of the 40th International Conference on Machine Learning}, ICML'23. JMLR.org.

\bibitem[{Albalak et~al.(2024)Albalak, Elazar, Xie, Longpre, Lambert, Wang, Muennighoff, Hou, Pan, Jeong et~al.}]{albalak2024survey}
Alon Albalak, Yanai Elazar, Sang~Michael Xie, Shayne Longpre, Nathan Lambert, Xinyi Wang, Niklas Muennighoff, Bairu Hou, Liangming Pan, Haewon Jeong, et~al. 2024.
\newblock A survey on data selection for language models.
\newblock \emph{arXiv preprint arXiv:2402.16827}.

\bibitem[{Besiroglu et~al.(2024)Besiroglu, Erdil, Barnett, and You}]{besiroglu2024chinchilla}
Tamay Besiroglu, Ege Erdil, Matthew Barnett, and Josh You. 2024.
\newblock Chinchilla scaling: A replication attempt.
\newblock \emph{arXiv preprint arXiv:2404.10102}.

\bibitem[{Bisk et~al.(2020)Bisk, Zellers, Gao, Choi et~al.}]{bisk2020piqa}
Yonatan Bisk, Rowan Zellers, Jianfeng Gao, Yejin Choi, et~al. 2020.
\newblock Piqa: Reasoning about physical commonsense in natural language.
\newblock In \emph{Proceedings of the AAAI conference on artificial intelligence}, volume~34, pages 7432--7439.

\bibitem[{Chung et~al.(2023)Chung, Kamar, and Amershi}]{chung2023increasing}
John Chung, Ece Kamar, and Saleema Amershi. 2023.
\newblock Increasing diversity while maintaining accuracy: Text data generation with large language models and human interventions.
\newblock In \emph{Proceedings of the 61st Annual Meeting of the Association for Computational Linguistics (Volume 1: Long Papers)}, pages 575--593.

\bibitem[{Clark et~al.(2019)Clark, Lee, Chang, Kwiatkowski, Collins, and Toutanova}]{clark2019boolq}
Christopher Clark, Kenton Lee, Ming-Wei Chang, Tom Kwiatkowski, Michael Collins, and Kristina Toutanova. 2019.
\newblock Boolq: Exploring the surprising difficulty of natural yes/no questions.
\newblock \emph{arXiv preprint arXiv:1905.10044}.

\bibitem[{Clark et~al.(2018)Clark, Cowhey, Etzioni, Khot, Sabharwal, Schoenick, and Tafjord}]{clark2018arc}
Peter Clark, Isaac Cowhey, Oren Etzioni, Tushar Khot, Ashish Sabharwal, Carissa Schoenick, and Oyvind Tafjord. 2018.
\newblock Think you have solved question answering? try arc, the ai2 reasoning challenge.
\newblock \emph{arXiv preprint arXiv:1803.05457}.

\bibitem[{Covington and McFall(2010)}]{Covington2010CuttingTG}
Michael~A. Covington and Joe~D. McFall. 2010.
\newblock \href {https://api.semanticscholar.org/CorpusID:18924254} {Cutting the gordian knot: The moving-average type–token ratio (mattr)}.
\newblock \emph{Journal of Quantitative Linguistics}, 17:100 -- 94.

\bibitem[{Ding et~al.(2024)Ding, Peng, Min, Kaiser, Yang, and Ray}]{ding2024semcoder}
Yangruibo Ding, Jinjun Peng, Marcus~J Min, Gail Kaiser, Junfeng Yang, and Baishakhi Ray. 2024.
\newblock Semcoder: Training code language models with comprehensive semantics.
\newblock \emph{arXiv preprint arXiv:2406.01006}.

\bibitem[{Gailly and Adler(1992)}]{gailly1992gnu}
Jean-loup Gailly and Mark Adler. 1992.
\newblock Gnu gzip.
\newblock \emph{GNU Operating System}.

\bibitem[{Goyal et~al.(2024)Goyal, Maini, Lipton, Raghunathan, and Kolter}]{goyal2024scaling}
Sachin Goyal, Pratyush Maini, Zachary~C Lipton, Aditi Raghunathan, and J~Zico Kolter. 2024.
\newblock Scaling laws for data filtering--data curation cannot be compute agnostic.
\newblock In \emph{Proceedings of the IEEE/CVF Conference on Computer Vision and Pattern Recognition}, pages 22702--22711.

\bibitem[{Hestness et~al.(2017)Hestness, Narang, Ardalani, Diamos, Jun, Kianinejad, Patwary, Yang, and Zhou}]{hestness2017deeplearningscalingpredictable}
Joel Hestness, Sharan Narang, Newsha Ardalani, Gregory Diamos, Heewoo Jun, Hassan Kianinejad, Md. Mostofa~Ali Patwary, Yang Yang, and Yanqi Zhou. 2017.
\newblock \href {http://arxiv.org/abs/1712.00409} {Deep learning scaling is predictable, empirically}.

\bibitem[{Hoffmann et~al.(2022)Hoffmann, Borgeaud, Mensch, Buchatskaya, Cai, Rutherford, de~Las~Casas, Hendricks, Welbl, Clark, Hennigan, Noland, Millican, van~den Driessche, Damoc, Guy, Osindero, Simonyan, Elsen, Rae, Vinyals, and Sifre}]{hoffmann2022chinchilla}
Jordan Hoffmann, Sebastian Borgeaud, Arthur Mensch, Elena Buchatskaya, Trevor Cai, Eliza Rutherford, Diego de~Las~Casas, Lisa~Anne Hendricks, Johannes Welbl, Aidan Clark, Tom Hennigan, Eric Noland, Katie Millican, George van~den Driessche, Bogdan Damoc, Aurelia Guy, Simon Osindero, Karen Simonyan, Erich Elsen, Jack~W. Rae, Oriol Vinyals, and Laurent Sifre. 2022.
\newblock An empirical analysis of compute-optimal large language model training.
\newblock In \emph{Advances in Neural Information Processing Systems (NeurIPS)}.

\bibitem[{John and Draper(1975)}]{doi:10.1080/00401706.1975.10489266}
R.~C.~St. John and N.~R. Draper. 1975.
\newblock \href {https://doi.org/10.1080/00401706.1975.10489266} {D-optimality for regression designs: A review}.
\newblock \emph{Technometrics}, 17(1):15--23.

\bibitem[{Kaplan et~al.(2020)Kaplan, McCandlish, Henighan, Brown, Chess, Child, Gray, Radford, Wu, and Amodei}]{kaplan2020scaling}
Jared Kaplan, Sam McCandlish, Tom Henighan, Tom~B Brown, Benjamin Chess, Rewon Child, Scott Gray, Alec Radford, Jeffrey Wu, and Dario Amodei. 2020.
\newblock \href {https://arxiv.org/abs/2001.08361} {Scaling laws for neural language models}.
\newblock \emph{arXiv preprint arXiv:2001.08361}.

\bibitem[{Lee et~al.(2022)Lee, Ippolito, Nystrom, Zhang, Eck, Callison-Burch, and Carlini}]{lee2022deduplicating}
Katherine Lee, Daphne Ippolito, Andrew Nystrom, Chiyuan Zhang, Douglas Eck, Chris Callison-Burch, and Nicholas Carlini. 2022.
\newblock Deduplicating training data makes language models better.
\newblock In \emph{Proceedings of the 60th Annual Meeting of the Association for Computational Linguistics (Volume 1: Long Papers)}, pages 8424--8445.

\bibitem[{Li et~al.(2016)Li, Galley, Brockett, Gao, and Dolan}]{li-etal-2016-diversity}
Jiwei Li, Michel Galley, Chris Brockett, Jianfeng Gao, and Bill Dolan. 2016.
\newblock \href {https://doi.org/10.18653/v1/N16-1014} {A diversity-promoting objective function for neural conversation models}.
\newblock In \emph{Proceedings of the 2016 Conference of the North {A}merican Chapter of the Association for Computational Linguistics: Human Language Technologies}, pages 110--119, San Diego, California. Association for Computational Linguistics.

\bibitem[{Li et~al.(2023)Li, Bubeck, Eldan, Giorno, Gunasekar, and Lee}]{li2023textbooksneediiphi15}
Yuanzhi Li, Sébastien Bubeck, Ronen Eldan, Allie~Del Giorno, Suriya Gunasekar, and Yin~Tat Lee. 2023.
\newblock \href {http://arxiv.org/abs/2309.05463} {Textbooks are all you need ii: phi-1.5 technical report}.

\bibitem[{Liu et~al.(2024)Liu, Wei, Liu, Si, Zhang, Rao, Zheng, Peng, Yang, Zhou et~al.}]{liu2024best}
Ruibo Liu, Jerry Wei, Fangyu Liu, Chenglei Si, Yanzhe Zhang, Jinmeng Rao, Steven Zheng, Daiyi Peng, Diyi Yang, Denny Zhou, et~al. 2024.
\newblock Best practices and lessons learned on synthetic data for language models.
\newblock \emph{arXiv preprint arXiv:2404.07503}.

\bibitem[{Long et~al.(2024)Long, Wang, Xiao, Zhao, Ding, Chen, and Wang}]{long2024llms}
Lin Long, Rui Wang, Ruixuan Xiao, Junbo Zhao, Xiao Ding, Gang Chen, and Haobo Wang. 2024.
\newblock On llms-driven synthetic data generation, curation, and evaluation: A survey.
\newblock \emph{arXiv preprint arXiv:2406.15126}.

\bibitem[{Maini et~al.(2024)Maini, Seto, Bai, Grangier, Zhang, and Jaitly}]{maini2024rephrasingwebrecipecompute}
Pratyush Maini, Skyler Seto, He~Bai, David Grangier, Yizhe Zhang, and Navdeep Jaitly. 2024.
\newblock \href {http://arxiv.org/abs/2401.16380} {Rephrasing the web: A recipe for compute and data-efficient language modeling}.

\bibitem[{Meister et~al.(2023)Meister, Pimentel, Wiher, and Cotterell}]{meister-etal-2023-locally}
Clara Meister, Tiago Pimentel, Gian Wiher, and Ryan Cotterell. 2023.
\newblock \href {https://doi.org/10.1162/tacl_a_00536} {Locally typical sampling}.
\newblock \emph{Transactions of the Association for Computational Linguistics}, 11:102--121.

\bibitem[{Mihaylov et~al.(2018)Mihaylov, Clark, Khot, and Sabharwal}]{mihaylov2018obqa}
Todor Mihaylov, Peter Clark, Tushar Khot, and Ashish Sabharwal. 2018.
\newblock Can a suit of armor conduct electricity? a new dataset for open book question answering.
\newblock \emph{arXiv preprint arXiv:1809.02789}.

\bibitem[{Muennighoff et~al.(2024)Muennighoff, Rush, Barak, Le~Scao, Tazi, Piktus, Pyysalo, Wolf, and Raffel}]{muennighoff2024scaling}
Niklas Muennighoff, Alexander Rush, Boaz Barak, Teven Le~Scao, Nouamane Tazi, Aleksandra Piktus, Sampo Pyysalo, Thomas Wolf, and Colin~A Raffel. 2024.
\newblock Scaling data-constrained language models.
\newblock \emph{Advances in Neural Information Processing Systems}, 36.

\bibitem[{Murphy(2012)}]{10.5555/2380985}
Kevin~P. Murphy. 2012.
\newblock \emph{Machine Learning: A Probabilistic Perspective}.
\newblock The MIT Press.

\bibitem[{Narayan et~al.(2024)Narayan, Chen, Bhatia, and Re}]{narayan2024cookbook}
Avanika Narayan, Mayee~F Chen, Kush Bhatia, and Christopher Re. 2024.
\newblock \href {https://openreview.net/forum?id=tbJXkDiiED} {Cookbook: A framework for improving {LLM} generative abilities via programmatic data generating templates}.
\newblock In \emph{ICLR 2024 Workshop on Navigating and Addressing Data Problems for Foundation Models}.

\bibitem[{Padmakumar et~al.(2023)Padmakumar, Hedayatnia, Jin, Lange, Kim, Peng, Liu, and Hakkani-Tur}]{padmakumar-etal-2023-investigating}
Vishakh Padmakumar, Behnam Hedayatnia, Di~Jin, Patrick Lange, Seokhwan Kim, Nanyun Peng, Yang Liu, and Dilek Hakkani-Tur. 2023.
\newblock \href {https://doi.org/10.18653/v1/2023.sigdial-1.50} {Investigating the representation of open domain dialogue context for transformer models}.
\newblock In \emph{Proceedings of the 24th Annual Meeting of the Special Interest Group on Discourse and Dialogue}, pages 538--547, Prague, Czechia. Association for Computational Linguistics.

\bibitem[{Penedo et~al.(2024)Penedo, Kydl{\'\i}{\v{c}}ek, Lozhkov, Mitchell, Raffel, Von~Werra, Wolf et~al.}]{penedo2024fineweb}
Guilherme Penedo, Hynek Kydl{\'\i}{\v{c}}ek, Anton Lozhkov, Margaret Mitchell, Colin Raffel, Leandro Von~Werra, Thomas Wolf, et~al. 2024.
\newblock The fineweb datasets: Decanting the web for the finest text data at scale.
\newblock \emph{arXiv preprint arXiv:2406.17557}.

\bibitem[{Penedo et~al.(2023)Penedo, Malartic, Hesslow, Cojocaru, Alobeidli, Cappelli, Pannier, Almazrouei, and Launay}]{penedo2023refinedweb}
Guilherme Penedo, Quentin Malartic, Daniel Hesslow, Ruxandra Cojocaru, Hamza Alobeidli, Alessandro Cappelli, Baptiste Pannier, Ebtesam Almazrouei, and Julien Launay. 2023.
\newblock \href {https://proceedings.neurips.cc/paper_files/paper/2023/file/fa3ed726cc5073b9c31e3e49a807789c-Paper-Datasets_and_Benchmarks.pdf} {The refinedweb dataset for falcon llm: Outperforming curated corpora with web data only}.
\newblock In \emph{Advances in Neural Information Processing Systems}, volume~36, pages 79155--79172. Curran Associates, Inc.

\bibitem[{Sakaguchi et~al.(2021)Sakaguchi, Bras, Bhagavatula, and Choi}]{sakaguchi2021winogrande}
Keisuke Sakaguchi, Ronan~Le Bras, Chandra Bhagavatula, and Yejin Choi. 2021.
\newblock Winogrande: An adversarial winograd schema challenge at scale.
\newblock \emph{Communications of the ACM}, 64(9):99--106.

\bibitem[{Salkar et~al.(2022)Salkar, Trikalinos, Wallace, and Nenkova}]{salkar2022self}
Nikita Salkar, Thomas Trikalinos, Byron~C Wallace, and Ani Nenkova. 2022.
\newblock \href {https://aclanthology.org/2022.aacl-short.42} {Self-repetition in abstractive neural summarizers}.
\newblock In \emph{Proceedings of the Conference of the Asia-Pacific Chapter of the Association for Computational Linguistics and the International Joint Conference on Natural Language Processing (AACL)}, volume 2022, pages 341--350.

\bibitem[{Sap et~al.(2019)Sap, Rashkin, Chen, LeBras, and Choi}]{sap2019siqa}
Maarten Sap, Hannah Rashkin, Derek Chen, Ronan LeBras, and Yejin Choi. 2019.
\newblock Socialiqa: Commonsense reasoning about social interactions.
\newblock \emph{arXiv preprint arXiv:1904.09728}.

\bibitem[{Shaib et~al.(2024)Shaib, Barrow, Sun, Siu, Wallace, and Nenkova}]{shaib2024standardizing}
Chantal Shaib, Joe Barrow, Jiuding Sun, Alexa~F Siu, Byron~C Wallace, and Ani Nenkova. 2024.
\newblock Standardizing the measurement of text diversity: A tool and a comparative analysis of scores.
\newblock \emph{CoRR}.

\bibitem[{Sorscher et~al.(2022)Sorscher, Geirhos, Shekhar, Ganguli, and Morcos}]{sorscher2022beyond}
Ben Sorscher, Robert Geirhos, Shashank Shekhar, Surya Ganguli, and Ari~S. Morcos. 2022.
\newblock \href {https://openreview.net/forum?id=UmvSlP-PyV} {Beyond neural scaling laws: beating power law scaling via data pruning}.
\newblock In \emph{Advances in Neural Information Processing Systems}.

\bibitem[{Tirumala et~al.(2024)Tirumala, Simig, Aghajanyan, and Morcos}]{tirumala2024d4}
Kushal Tirumala, Daniel Simig, Armen Aghajanyan, and Ari Morcos. 2024.
\newblock D4: Improving llm pretraining via document de-duplication and diversification.
\newblock \emph{Advances in Neural Information Processing Systems}, 36.

\bibitem[{Touvron et~al.(2023)Touvron, Lavril, Izacard, Martinet, Lachaux, Lacroix, Rozi{\`e}re, Goyal, Hambro, Azhar et~al.}]{touvron2023llama}
Hugo Touvron, Thibaut Lavril, Gautier Izacard, Xavier Martinet, Marie-Anne Lachaux, Timoth{\'e}e Lacroix, Baptiste Rozi{\`e}re, Naman Goyal, Eric Hambro, Faisal Azhar, et~al. 2023.
\newblock Llama: Open and efficient foundation language models.
\newblock \emph{arXiv preprint arXiv:2302.13971}.

\bibitem[{Wang et~al.(2024)Wang, Zhang, Du, Zhang, and Chu}]{wang2024survey}
Jiahao Wang, Bolin Zhang, Qianlong Du, Jiajun Zhang, and Dianhui Chu. 2024.
\newblock A survey on data selection for llm instruction tuning.
\newblock \emph{arXiv preprint arXiv:2402.05123}.

\bibitem[{Wang et~al.(2018)Wang, Zhu, Torralba, and Efros}]{wang2018dataset}
Tongzhou Wang, Jun-Yan Zhu, Antonio Torralba, and Alexei~A Efros. 2018.
\newblock Dataset distillation.
\newblock \emph{arXiv preprint arXiv:1811.10959}.

\bibitem[{Xie et~al.(2023)Xie, Santurkar, Ma, and Liang}]{xie2023data}
Sang~Michael Xie, Shibani Santurkar, Tengyu Ma, and Percy Liang. 2023.
\newblock \href {https://openreview.net/forum?id=uPSQv0leAu} {Data selection for language models via importance resampling}.
\newblock In \emph{Thirty-seventh Conference on Neural Information Processing Systems}.

\bibitem[{Zellers et~al.(2019)Zellers, Holtzman, Bisk, Farhadi, and Choi}]{zellers2019hellaswag}
Rowan Zellers, Ari Holtzman, Yonatan Bisk, Ali Farhadi, and Yejin Choi. 2019.
\newblock Hellaswag: Can a machine really finish your sentence?
\newblock \emph{arXiv preprint arXiv:1905.07830}.

\bibitem[{Zhao et~al.(2023)Zhao, Zhou, Li, Tang, Wang, Hou, Min, Zhang, Zhang, Dong et~al.}]{zhao2023survey}
Wayne~Xin Zhao, Kun Zhou, Junyi Li, Tianyi Tang, Xiaolei Wang, Yupeng Hou, Yingqian Min, Beichen Zhang, Junjie Zhang, Zican Dong, et~al. 2023.
\newblock A survey of large language models.
\newblock \emph{arXiv preprint arXiv:2303.18223}.

\end{thebibliography}
\bibliographystyle{acl_natbib}

\clearpage

\onecolumn

\appendix

\section{Details of Data Synthesis}

Here we provide the instruction prompt that is used for data synthesis, which is used to rewrite with a Llama-3-70B-instruct (\url{https://ai.meta.com/blog/meta-llama-3/}) model to rewrite provided documents from the pretraining data. 
The data for synthesis was sourced from a directory with JSONL files organized by group numbers and shards, and the model was configured to process sequences up to 8196 tokens in length. Computational precision was optimized for specific hardware by enabling BF16 and disabling FP16, with a batch size of 8 per device to ensure efficient processing and resource utilization. We provide the instruction prompt here:

\textit{{Create a common sense reasoning problem-answer pair based on the following text.} However, if it's impossible to create a problem, rewrite the text to be a textbook style language that is clear and concise. Only provide the relevant response and do not say anything else. Do not assume the reader to know anything about the text, so make sure to provide the context for the reasoning problem.}

\bigskip
\noindent \textit{{Text:} \\
\{Pretraining Document}\}
\bigskip \\
\noindent \textit{{Response:} \\
}

\section{Details of Data Selection}

We employ data selection as described in~\citet{xie2023data}. 
Here we provide additional details into the feature extraction process from documents. 
Due to memory limitations on our computational resources, we divided the RefinedWeb dataset into 16 distinct shards. From each shard, we selectively sampled a subset of data tailored to our target specifications.
The entire sampling process typically requires approximately 1.5 days to complete across all methodologies. It is important to note that variations in the tokenizer's vocabulary do not significantly affect the sampling speed. This observation suggests that the vocabulary size primarily influences the sentence compression ratio rather than the processing time.

\section{Computing Text Syntheticity}

To accurately assess syntheticity, it is essential to compute the perplexity for each document. This involves deploying a language model with a context length of $1024$ tokens to process all documents. The average perplexity score across these documents serves as the metric for syntheticity.

Given the computationally demanding nature of calculating perplexity with language models, we strategically sampled $25\%$ of complete documents from each dataset. This sampling strategy results in a substantial volume of data, ranging from approximately $100$ million to several billion subword tokens, ensuring a robust and efficient analysis.

\section{Scaling Law Constant Estimation}
In this work, we introduce a scaling law for language modeling systems, defined as $\hat{G}(N,D) = E + \frac{A}{N^\alpha} + \frac{B}{D^\beta}$. Here, $\hat{G}(N,D)$ estimates accuracy, with $N$ as model size and $D$ as dataset size. Constants $E$, $A$, $\alpha$, $B$, $\beta$, $c1$, and $c2$ are parameters to be determined.

The estimation of this scaling law constants involved analyzing a dataset of 210 data points, each representing different model and dataset sizes with corresponding training losses and accuracy scores. 
These estimation accounted for the refinement of the training data that incorporate additional factors such as diversity and syntheticity into the dataset size. 
Further, different transformations of the dataset size were included to determine how these factors could be integrated effectively. 
The accuracy of the model was then obtained for each of these refinements.
This comprehensive dataset allowed for robust parameter estimation.
Parameter estimation was achieved through nonlinear curve fitting, aiming to align the scaling law's predictions with observed training losses. The process included:
\begin{enumerate}
    \item \textbf{Model Definition:} Formulating the scaling law as a function with parameters to estimate. Overall, we have experimented with four equations for $D_q$:
\begin{table}[h]
\centering
\begin{tabular}{c|c}
\textbf{Equation} & \textbf{$R^2$} \\ \hline
$D \cdot \exp(c_1 \cdot \text{Dr}(D) + c_2 \cdot \text{S}(D))$ & 0.45 \\ \hline
$D \cdot \text{Dr}(D)^{c_1} \cdot \exp(c_2 \cdot \text{S}(D))$ & 0.23 \\ \hline
$D \cdot \exp(c_1 \cdot \text{Dr}(D)) \cdot \text{S}(D)^{c_2}$ & 0.19 \\ \hline
$D \cdot \text{Dr}(D)^{c_1} \cdot \text{S}(D)^{c_2}$ & 0.35 \\ 
\end{tabular}
\caption{Equations and their corresponding $R^2$ values}
\label{tab:equations}
\end{table}
    \item \textbf{Initial Guesses:} Setting initial parameter values based on \citet{besiroglu2024chinchilla}. Initial guesses were $E = 1.8172$, $A = 482.01$, $\alpha = 0.3478$, $B = 2085.43$, $\beta = 0.3658$, and we proposed to set $c1 = 0.5$, and $c2 = 0.5$.
    \item \textbf{Optimization Algorithm:} Utilizing the 'curve\_fit' function from 'scipy.optimize' to perform non-linear least squares fitting. The algorithm adjusted the parameters to minimize the sum of the squares of the differences between observed and predicted values.
    \item \textbf{Convergence and Validation:} Iterating the fitting process until parameter changes minimized, and validating the model by examining residuals and fit quality. The process ensured that the parameters converged effectively, representing the trends in the data accurately.
\end{enumerate}

During curve fitting, the goodness of fit was assessed using the R-squared value, which measures the proportion of variance in the observed data that is predictable from the model inputs. 
This iterative process of refinement and evaluation helped in achieving the best possible fit between the predicted and observed accuracies, enhancing the scaling law's ability to predict training losses across various settings.
We stop the iteration at 200.

This process refined the estimates of $E$, $A$, $\alpha$, $B$, $\beta$, $c1$, and $c2$, enhancing the scaling law's ability to predict training losses across various settings, thus supporting efficient resource allocation and model design in language modeling. 
The refined constants provided a more accurate description of how training loss scales with changes in model size and dataset size, incorporating the effects of diversity and syntheticity through $c1$ and $c2$.

\section{Deriving Effective Token $D_q$ Equation}\label{appendix:inverse}

We derive the formula to obtain the number of \textit{effective tokens} as a function of the loss.

Original formula:
\begin{equation}
\hat L(N,D) \triangleq E + \frac{A}{N^\alpha} + \frac{B}{D_{\text{q}}^\beta}
\end{equation}

We consider shorten the loss $\hat L(N,D)$ as $L$.
\begin{equation}
L \triangleq E + \frac{A}{N^\alpha} + \frac{B}{D_{\text{q}}^\beta}
\end{equation}

Move the E to the left:
\begin{equation}
L - E - \frac{A}{N^\alpha} \triangleq \frac{B}{D_{\text{q}}^\beta}
\end{equation}

Make same denominator:
\begin{equation}
 \frac{L N^\alpha - E N^\alpha - A}{N^\alpha} \triangleq \frac{B}{D_{\text{q}}^\beta}
\end{equation}

Group the $N^\alpha$:
\begin{equation}
 \frac{(L - E) N^\alpha - A}{N^\alpha} \triangleq \frac{B}{D_{\text{q}}^\beta}
\end{equation}

Flip Both:
\begin{equation}
 \frac{N^\alpha}{(L - E) N^\alpha - A} \triangleq \frac{D_{\text{q}}^\beta}{B}
\end{equation}

Isolate D to the beta on the right:
\begin{equation}
 \frac{B N^\alpha}{(L - E) N^\alpha - A} \triangleq D_{\text{q}}^\beta
\end{equation}

Apply root of beta to get D effective tokens

\begin{equation}
D_{\text{q}} \triangleq \left(\frac{B N^{\alpha}}{(L - E) N^\alpha - A}\right)^{1/\beta}
\end{equation}

Here we provide additional details regarding the process of feature extraction from documents.
Due to the memory constraints on the machines, we split the RefinedWeb data into 16 shards, and sampled a subset from each shard based on the target data. 
This process takes around 1.5 days on average for all approaches, meaning that the change in tokenizer's vocabulary does not result in noticeable differences in sampling speed, since vocabulary also defines sentence compression ratio.

\section{Diversity and syntheticity Result Table}
\centering

\resizebox{0.6\columnwidth}{!}{
\begin{tabular}{llrll}
\toprule
 & Data & \%  & Diversity & syntheticity \\
\midrule
1 & Random & 10 & 0.37750 & 0.02699 \\
2 & Random & 20 & 0.37783 & 0.02682 \\
3 & Random & 30 & 0.37833 & 0.02675 \\
4 & Random & 40 & 0.37853 & 0.02705 \\
5 & Random & 50 & 0.38348 & 0.02661 \\
6 & Random & 60 & 0.38003 & 0.02658 \\
7 & Random & 70 & 0.38618 & 0.02656 \\
8 & Random & 80 & 0.42511 & 0.02649 \\
9 & Random & 90 & 0.46301 & 0.02642 \\
10 & Random & 100 & 0.36370 & 0.02635 \\
11 & Selection & 10 & 0.36187 & 0.04230 \\
12 & Selection & 20 & 0.36189 & 0.04080 \\
13 & Selection & 30 & 0.36186 & 0.04102 \\
14 & Selection & 40 & 0.36186 & 0.04069 \\
15 & Selection & 50 & 0.36187 & 0.04102 \\
16 & Selection & 60 & 0.36188 & 0.04089 \\
17 & Selection & 70 & 0.36189 & 0.04065 \\
18 & Selection & 80 & 0.36189 & 0.04015 \\
19 & Selection & 90 & 0.36190 & 0.04003 \\
20 & Selection & 100 & 0.29054 & 0.03990 \\
21 & Selection + Synthesis & 10 & 0.28586 & 0.13058 \\
22 & Selection + Synthesis & 20 & 0.28585 & 0.11919 \\
23 & Selection + Synthesis & 30 & 0.28584 & 0.12308 \\
24 & Selection + Synthesis & 40 & 0.28579 & 0.12383 \\
25 & Selection + Synthesis & 50 & 0.28580 & 0.12489 \\
26 & Selection + Synthesis & 60 & 0.28577 & 0.12719 \\
27 & Selection + Synthesis & 70 & 0.28579 & 0.13113 \\
28 & Selection + Synthesis & 80 & 0.28581 & 0.12656 \\
29 & Selection + Synthesis & 90 & 0.28578 & 0.12002 \\
30 & Selection + Synthesis & 100 & 0.28578 & 0.11902 \\
\bottomrule
\end{tabular}}

\newpage

\section{Scaling Law Result Table}
\centering

\resizebox{0.8\columnwidth}{!}{\begin{tabular}{lrlrllll}
\toprule
 & Size (M) & Data & \%  & N. Tokens & Train Loss & Eval Loss & Avg. Acc. \\
\midrule
1 & 25 & Random & 10 & 1,083,200,970 & 1.36 & 6.89 & 37.87 \\
2 & 50 & Random & 10 & 1,083,200,970 & 3.26 & 4.00 & 35.30 \\
3 & 75 & Random & 10 & 1,083,200,970 & 2.77 & 3.62 & 38.93 \\
4 & 125 & Random & 10 & 1,083,200,970 & 2.69 & 3.58 & 39.31 \\
5 & 500 & Random & 10 & 1,083,200,970 & 1.78 & 4.51 & 40.47 \\
6 & 1500 & Random & 10 & 1,083,200,970 & 0.25 & 11.33 & 40.68 \\
7 & 25 & Random & 20 & 2,178,049,311 & 1.42 & 5.60 & 40.76 \\
8 & 50 & Random & 20 & 2,178,049,311 & 3.28 & 3.97 & 37.43 \\
9 & 75 & Random & 20 & 2,178,049,311 & 2.81 & 3.51 & 39.06 \\
10 & 125 & Random & 20 & 2,178,049,311 & 2.70 & 3.45 & 40.04 \\
11 & 350 & Random & 20 & 2,178,049,311 & 2.35 & 3.37 & 41.59 \\
12 & 500 & Random & 20 & 2,178,049,311 & 2.18 & 3.43 & 43.29 \\
13 & 1500 & Random & 20 & 2,178,049,311 & 1.29 & 5.10 & 42.46 \\
14 & 25 & Random & 30 & 3,301,058,727 & 3.14 & 3.82 & 38.30 \\
15 & 50 & Random & 30 & 3,301,058,727 & 3.29 & 3.99 & 37.56 \\
16 & 75 & Random & 30 & 3,301,058,727 & 2.82 & 3.50 & 39.66 \\
17 & 125 & Random & 30 & 3,301,058,727 & 2.71 & 3.38 & 40.47 \\
18 & 350 & Random & 30 & 3,301,058,727 & 2.41 & 3.23 & 42.11 \\
19 & 500 & Random & 30 & 3,301,058,727 & 2.30 & 3.21 & 43.12 \\
20 & 1500 & Random & 30 & 3,301,058,727 & 1.70 & 3.53 & 45.33 \\
21 & 25 & Random & 40 & 4,391,680,343 & 3.15 & 3.82 & 37.88 \\
22 & 50 & Random & 40 & 4,391,680,343 & 3.28 & 3.98 & 36.27 \\
23 & 75 & Random & 40 & 4,391,680,343 & 2.83 & 3.48 & 38.96 \\
24 & 125 & Random & 40 & 4,391,680,343 & 2.72 & 3.40 & 41.05 \\
25 & 350 & Random & 40 & 4,391,680,343 & 2.44 & 3.16 & 43.36 \\
26 & 500 & Random & 40 & 4,391,680,343 & 2.32 & 3.12 & 43.50 \\
27 & 1500 & Random & 40 & 4,391,680,343 & 2.01 & 3.12 & 45.19 \\
28 & 25 & Random & 50 & 5,471,561,263 & 3.15 & 3.85 & 37.80 \\
29 & 50 & Random & 50 & 5,471,561,263 & 3.28 & 3.98 & 36.51 \\
30 & 75 & Random & 50 & 5,471,561,263 & 2.91 & 3.53 & 40.07 \\
31 & 125 & Random & 50 & 5,471,561,263 & 2.73 & 3.38 & 39.82 \\
32 & 350 & Random & 50 & 5,471,561,263 & 2.46 & 3.14 & 42.90 \\
33 & 500 & Random & 50 & 5,471,561,263 & 2.36 & 3.06 & 43.56 \\
34 & 1500 & Random & 50 & 5,471,561,263 & 2.11 & 3.02 & 46.22 \\
35 & 25 & Random & 60 & 6,599,971,622 & 3.16 & 3.84 & 37.78 \\
36 & 50 & Random & 60 & 6,599,971,622 & 3.29 & 3.98 & 35.82 \\
37 & 75 & Random & 60 & 6,599,971,622 & 2.84 & 3.49 & 39.25 \\
38 & 125 & Random & 60 & 6,599,971,622 & 2.72 & 3.34 & 40.81 \\
39 & 350 & Random & 60 & 6,599,971,622 & 2.46 & 3.10 & 43.35 \\
40 & 500 & Random & 60 & 6,599,971,622 & 2.51 & 3.10 & 43.94 \\
41 & 1500 & Random & 60 & 6,599,971,622 & 2.16 & 2.92 & 46.82 \\
42 & 25 & Random & 70 & 7,688,714,499 & 3.15 & 3.83 & 38.24 \\
43 & 50 & Random & 70 & 7,688,714,499 & 3.28 & 3.97 & 37.20 \\
44 & 75 & Random & 70 & 7,688,714,499 & 2.85 & 3.49 & 38.70 \\
45 & 125 & Random & 70 & 7,688,714,499 & 2.76 & 3.38 & 40.35 \\
\bottomrule
\end{tabular}}

\resizebox{0.8\columnwidth}{!}{\begin{tabular}{lrlrllll}
\toprule
 & Size (M) & Data & \%  & N. Tokens & Train Loss & Eval Loss & Avg. Acc. \\
\midrule
46 & 350 & Random & 70 & 7,688,714,499 & 2.47 & 3.12 & 43.69 \\
47 & 500 & Random & 70 & 7,688,714,499 & 2.52 & 3.09 & 43.50 \\
48 & 1500 & Random & 70 & 7,688,714,499 & 2.19 & 2.91 & 47.80 \\
49 & 25 & Random & 80 & 8,761,608,715 & 3.14 & 3.81 & 38.01 \\
50 & 50 & Random & 80 & 8,761,608,715 & 3.29 & 3.96 & 37.44 \\
51 & 75 & Random & 80 & 8,761,608,715 & 2.85 & 3.49 & 39.55 \\
52 & 125 & Random & 80 & 8,761,608,715 & 2.74 & 3.39 & 40.85 \\
53 & 350 & Random & 80 & 8,761,608,715 & 2.48 & 3.09 & 43.89 \\
54 & 500 & Random & 80 & 8,761,608,715 & 2.40 & 3.02 & 44.63 \\
55 & 1500 & Random & 80 & 8,761,608,715 & 2.23 & 2.87 & 47.97 \\
56 & 25 & Random & 90 & 9,882,886,144 & 3.15 & 3.85 & 37.48 \\
57 & 50 & Random & 90 & 9,882,886,144 & 3.28 & 3.98 & 37.65 \\
58 & 75 & Random & 90 & 9,882,886,144 & 2.83 & 3.46 & 39.45 \\
59 & 125 & Random & 90 & 9,882,886,144 & 2.73 & 3.34 & 40.63 \\
60 & 350 & Random & 90 & 9,882,886,144 & 2.47 & 3.08 & 43.39 \\
61 & 500 & Random & 90 & 9,882,886,144 & 2.39 & 3.01 & 44.13 \\
62 & 1500 & Random & 90 & 9,882,886,144 & 2.23 & 2.85 & 49.16 \\
63 & 25 & Random & 100 & 10,993,147,242 & 3.15 & 3.84 & 38.27 \\
64 & 50 & Random & 100 & 10,993,147,242 & 3.29 & 3.97 & 36.44 \\
65 & 75 & Random & 100 & 10,993,147,242 & 2.84 & 3.46 & 38.73 \\
66 & 125 & Random & 100 & 10,993,147,242 & 2.73 & 3.34 & 40.85 \\
67 & 350 & Random & 100 & 10,993,147,242 & 2.49 & 3.09 & 43.81 \\
68 & 500 & Random & 100 & 10,993,147,242 & 2.41 & 2.98 & 45.09 \\
69 & 1500 & Random & 100 & 10,993,147,242 & 2.15 & 2.85 & 48.23 \\
70 & 25 & Selection & 10 & 708,363,509 & 2.67 & 4.70 & 39.02 \\
71 & 50 & Selection & 10 & 708,363,509 & 2.45 & 4.70 & 40.81 \\
72 & 75 & Selection & 10 & 708,363,509 & 2.29 & 4.79 & 39.75 \\
73 & 125 & Selection & 10 & 708,363,509 & 2.12 & 5.18 & 40.57 \\
74 & 350 & Selection & 10 & 708,363,509 & 1.37 & 7.71 & 41.13 \\
75 & 500 & Selection & 10 & 708,363,509 & 0.95 & 10.27 & 40.57 \\
76 & 1500 & Selection & 10 & 708,363,509 & 0.10 & 14.46 & 41.13 \\
77 & 25 & Selection & 20 & 1,417,265,043 & 2.68 & 4.65 & 39.20 \\
78 & 50 & Selection & 20 & 1,417,265,043 & 2.48 & 4.49 & 40.40 \\
79 & 75 & Selection & 20 & 1,417,265,043 & 2.33 & 4.35 & 41.44 \\
80 & 125 & Selection & 20 & 1,417,265,043 & 2.25 & 4.28 & 41.71 \\
81 & 350 & Selection & 20 & 1,417,265,043 & 1.81 & 4.91 & 43.07 \\
82 & 500 & Selection & 20 & 1,417,265,043 & 1.62 & 5.80 & 43.17 \\
83 & 1500 & Selection & 20 & 1,417,265,043 & 0.35 & 11.82 & 43.16 \\
84 & 25 & Selection & 30 & 2,127,218,639 & 2.68 & 4.65 & 39.51 \\
85 & 50 & Selection & 30 & 2,127,218,639 & 2.49 & 4.44 & 40.82 \\
86 & 75 & Selection & 30 & 2,127,218,639 & 2.35 & 4.31 & 41.64 \\
87 & 125 & Selection & 30 & 2,127,218,639 & 2.24 & 4.23 & 42.39 \\
88 & 500 & Selection & 30 & 2,127,218,639 & 1.80 & 4.58 & 44.37 \\
89 & 1500 & Selection & 30 & 2,127,218,639 & 0.83 & 7.70 & 43.12 \\
90 & 25 & Selection & 40 & 2,836,208,025 & 2.69 & 4.58 & 39.39 \\
\bottomrule
\end{tabular}}

\resizebox{0.8\columnwidth}{!}{\begin{tabular}{lrlrllll}
\toprule
 & Size (M) & Data & \%  & N. Tokens & Train Loss & Eval Loss & Avg. Acc. \\
\midrule
91 & 50 & Selection & 40 & 2,836,208,025 & 2.51 & 4.42 & 40.65 \\
92 & 75 & Selection & 40 & 2,836,208,025 & 2.35 & 4.25 & 40.97 \\
93 & 125 & Selection & 40 & 2,836,208,025 & 2.24 & 4.13 & 42.15 \\
94 & 350 & Selection & 40 & 2,836,208,025 & 1.96 & 4.09 & 44.44 \\
95 & 500 & Selection & 40 & 2,836,208,025 & 1.87 & 4.11 & 45.21 \\
96 & 1500 & Selection & 40 & 2,836,208,025 & 1.21 & 5.72 & 45.00 \\
97 & 25 & Selection & 50 & 3,544,568,369 & 2.67 & 4.57 & 38.82 \\
98 & 50 & Selection & 50 & 3,544,568,369 & 2.50 & 4.41 & 40.89 \\
99 & 75 & Selection & 50 & 3,544,568,369 & 2.37 & 4.25 & 41.55 \\
100 & 125 & Selection & 50 & 3,544,568,369 & 2.29 & 4.13 & 42.49 \\
101 & 350 & Selection & 50 & 3,544,568,369 & 2.01 & 3.94 & 45.30 \\
102 & 500 & Selection & 50 & 3,544,568,369 & 1.90 & 3.96 & 45.42 \\
103 & 1500 & Selection & 50 & 3,544,568,369 & 1.39 & 4.93 & 46.25 \\
104 & 25 & Selection & 60 & 4,253,350,223 & 2.66 & 4.57 & 40.01 \\
105 & 50 & Selection & 60 & 4,253,350,223 & 2.49 & 4.42 & 41.09 \\
106 & 75 & Selection & 60 & 4,253,350,223 & 2.36 & 4.22 & 41.41 \\
107 & 125 & Selection & 60 & 4,253,350,223 & 2.28 & 4.13 & 42.84 \\
108 & 350 & Selection & 60 & 4,253,350,223 & 2.00 & 3.93 & 44.87 \\
109 & 500 & Selection & 60 & 4,253,350,223 & 1.93 & 3.87 & 44.92 \\
110 & 1500 & Selection & 60 & 4,253,350,223 & 1.74 & 3.84 & 47.25 \\
111 & 25 & Selection & 70 & 4,962,280,568 & 2.67 & 4.61 & 39.34 \\
112 & 50 & Selection & 70 & 4,962,280,568 & 2.49 & 4.36 & 40.86 \\
113 & 75 & Selection & 70 & 4,962,280,568 & 2.42 & 4.24 & 42.50 \\
114 & 125 & Selection & 70 & 4,962,280,568 & 2.30 & 4.11 & 42.17 \\
115 & 350 & Selection & 70 & 4,962,280,568 & 2.01 & 3.86 & 45.09 \\
116 & 500 & Selection & 70 & 4,962,280,568 & 1.93 & 3.81 & 45.24 \\
117 & 1500 & Selection & 70 & 4,962,280,568 & 1.72 & 3.78 & 47.51 \\
118 & 25 & Selection & 80 & 5,670,003,836 & 2.67 & 4.60 & 39.64 \\
119 & 50 & Selection & 80 & 5,670,003,836 & 2.50 & 4.36 & 40.55 \\
120 & 75 & Selection & 80 & 5,670,003,836 & 2.37 & 4.19 & 41.86 \\
121 & 125 & Selection & 80 & 5,670,003,836 & 2.27 & 4.11 & 43.11 \\
122 & 350 & Selection & 80 & 5,670,003,836 & 2.02 & 3.86 & 44.84 \\
123 & 500 & Selection & 80 & 5,670,003,836 & 1.95 & 3.79 & 45.46 \\
124 & 1500 & Selection & 80 & 5,670,003,836 & 1.65 & 3.87 & 47.66 \\
125 & 25 & Selection & 90 & 6,378,582,091 & 2.68 & 4.60 & 39.57 \\
126 & 50 & Selection & 90 & 6,378,582,091 & 2.50 & 4.38 & 40.62 \\
127 & 75 & Selection & 90 & 6,378,582,091 & 2.35 & 4.18 & 41.34 \\
128 & 125 & Selection & 90 & 6,378,582,091 & 2.30 & 4.12 & 42.89 \\
129 & 350 & Selection & 90 & 6,378,582,091 & 2.02 & 3.84 & 44.78 \\
130 & 500 & Selection & 90 & 6,378,582,091 & 1.97 & 3.75 & 46.08 \\
131 & 1500 & Selection & 90 & 6,378,582,091 & 1.81 & 3.62 & 49.25 \\
132 & 25 & Selection & 100 & 7,087,328,618 & 2.68 & 4.60 & 39.49 \\
133 & 50 & Selection & 100 & 7,087,328,618 & 2.50 & 4.38 & 41.10 \\
134 & 75 & Selection & 100 & 7,087,328,618 & 2.36 & 4.22 & 41.86 \\
135 & 125 & Selection & 100 & 7,087,328,618 & 2.27 & 4.08 & 42.88 \\
\bottomrule
\end{tabular}}

\resizebox{0.8\columnwidth}{!}{\begin{tabular}{lrlrllll}
\toprule
 & Size (M) & Data & \%  & N. Tokens & Train Loss & Eval Loss & Avg. Acc. \\
\midrule
136 & 350 & Selection & 100 & 7,087,328,618 & 2.02 & 3.80 & 45.61 \\
137 & 500 & Selection & 100 & 7,087,328,618 & 1.96 & 3.75 & 46.51 \\
138 & 1500 & Selection & 100 & 7,087,328,618 & 1.68 & 3.74 & 48.82 \\
139 & 25 & Selection + Synthesis & 10 & 250,378,189 & 1.36 & 6.89 & 37.87 \\
140 & 50 & Selection + Synthesis & 10 & 250,378,189 & 1.49 & 6.15 & 38.25 \\
141 & 75 & Selection + Synthesis & 10 & 250,378,189 & 0.85 & 12.24 & 38.96 \\
142 & 125 & Selection + Synthesis & 10 & 250,378,189 & 0.49 & 15.56 & 38.55 \\
143 & 350 & Selection + Synthesis & 10 & 250,378,189 & 0.05 & 18.64 & 39.86 \\
144 & 500 & Selection + Synthesis & 10 & 250,378,189 & 0.03 & 17.01 & 38.89 \\
145 & 1500 & Selection + Synthesis & 10 & 250,378,189 & 0.02 & 13.44 & 40.32 \\
146 & 25 & Selection + Synthesis & 20 & 500,768,330 & 1.42 & 5.60 & 40.76 \\
147 & 50 & Selection + Synthesis & 20 & 500,768,330 & 1.52 & 5.51 & 37.67 \\
148 & 75 & Selection + Synthesis & 20 & 500,768,330 & 1.14 & 7.13 & 40.78 \\
149 & 125 & Selection + Synthesis & 20 & 500,768,330 & 0.94 & 8.89 & 40.08 \\
150 & 350 & Selection + Synthesis & 20 & 500,768,330 & 0.20 & 16.41 & 40.53 \\
151 & 500 & Selection + Synthesis & 20 & 500,768,330 & 0.08 & 17.22 & 40.58 \\
152 & 1500 & Selection + Synthesis & 20 & 500,768,330 & 0.03 & 14.28 & 41.80 \\
153 & 25 & Selection + Synthesis & 30 & 751,577,046 & 1.45 & 5.23 & 39.44 \\
154 & 50 & Selection + Synthesis & 30 & 751,577,046 & 1.54 & 5.36 & 38.48 \\
155 & 75 & Selection + Synthesis & 30 & 751,577,046 & 1.21 & 5.92 & 41.67 \\
156 & 125 & Selection + Synthesis & 30 & 751,577,046 & 1.08 & 6.74 & 41.88 \\
157 & 350 & Selection + Synthesis & 30 & 751,577,046 & 0.49 & 11.22 & 41.61 \\
158 & 500 & Selection + Synthesis & 30 & 751,577,046 & 0.26 & 14.23 & 41.97 \\
159 & 1500 & Selection + Synthesis & 30 & 751,577,046 & 0.04 & 14.48 & 42.49 \\
160 & 25 & Selection + Synthesis & 40 & 1,002,469,726 & 1.44 & 5.14 & 39.81 \\
161 & 50 & Selection + Synthesis & 40 & 1,002,469,726 & 1.58 & 5.25 & 38.54 \\
162 & 75 & Selection + Synthesis & 40 & 1,002,469,726 & 1.23 & 5.23 & 41.39 \\
163 & 125 & Selection + Synthesis & 40 & 1,002,469,726 & 1.13 & 5.88 & 41.33 \\
164 & 350 & Selection + Synthesis & 40 & 1,002,469,726 & 0.70 & 9.07 & 42.04 \\
165 & 500 & Selection + Synthesis & 40 & 1,002,469,726 & 0.48 & 10.96 & 43.47 \\
166 & 1500 & Selection + Synthesis & 40 & 1,002,469,726 & 0.07 & 13.62 & 43.42 \\
167 & 25 & Selection + Synthesis & 50 & 1,253,583,976 & 1.45 & 4.95 & 39.38 \\
168 & 50 & Selection + Synthesis & 50 & 1,253,583,976 & 1.54 & 5.23 & 38.74 \\
169 & 75 & Selection + Synthesis & 50 & 1,253,583,976 & 1.25 & 4.96 & 42.43 \\
170 & 125 & Selection + Synthesis & 50 & 1,253,583,976 & 1.17 & 5.29 & 42.77 \\
171 & 350 & Selection + Synthesis & 50 & 1,253,583,976 & 0.82 & 7.52 & 41.65 \\
172 & 500 & Selection + Synthesis & 50 & 1,253,583,976 & 0.64 & 9.17 & 43.37 \\
173 & 1500 & Selection + Synthesis & 50 & 1,253,583,976 & 0.15 & 12.36 & 43.18 \\
174 & 25 & Selection + Synthesis & 60 & 1,504,223,685 & 1.45 & 5.01 & 39.68 \\
175 & 50 & Selection + Synthesis & 60 & 1,504,223,685 & 1.54 & 5.11 & 37.69 \\
176 & 75 & Selection + Synthesis & 60 & 1,504,223,685 & 1.26 & 4.93 & 42.72 \\
177 & 125 & Selection + Synthesis & 60 & 1,504,223,685 & 1.18 & 5.00 & 43.10 \\
178 & 350 & Selection + Synthesis & 60 & 1,504,223,685 & 0.90 & 6.43 & 41.92 \\
179 & 500 & Selection + Synthesis & 60 & 1,504,223,685 & 0.75 & 7.65 & 42.99 \\
180 & 1500 & Selection + Synthesis & 60 & 1,504,223,685 & 0.21 & 11.04 & 44.03 \\
\bottomrule
\end{tabular}}

\resizebox{0.8\columnwidth}{!}{\begin{tabular}{lrlrllll}
\toprule
 & Size (M) & Data & \%  & N. Tokens & Train Loss & Eval Loss & Avg. Acc. \\
\midrule
181 & 25 & Selection + Synthesis & 70 & 1,754,577,326 & 1.46 & 4.99 & 40.42 \\
182 & 50 & Selection + Synthesis & 70 & 1,754,577,326 & 1.55 & 5.16 & 37.51 \\
183 & 75 & Selection + Synthesis & 70 & 1,754,577,326 & 1.27 & 4.81 & 42.51 \\
184 & 125 & Selection + Synthesis & 70 & 1,754,577,326 & 1.20 & 4.89 & 43.28 \\
185 & 350 & Selection + Synthesis & 70 & 1,754,577,326 & 0.95 & 6.18 & 43.47 \\
186 & 500 & Selection + Synthesis & 70 & 1,754,577,326 & 0.82 & 6.79 & 43.52 \\
187 & 1500 & Selection + Synthesis & 70 & 1,754,577,326 & 0.19 & 12.85 & 43.32 \\
188 & 25 & Selection + Synthesis & 80 & 2,004,994,693 & 1.46 & 5.03 & 40.48 \\
189 & 50 & Selection + Synthesis & 80 & 2,004,994,693 & 1.57 & 5.13 & 38.29 \\
190 & 75 & Selection + Synthesis & 80 & 2,004,994,693 & 1.27 & 4.70 & 42.92 \\
191 & 125 & Selection + Synthesis & 80 & 2,004,994,693 & 1.21 & 4.77 & 43.26 \\
192 & 350 & Selection + Synthesis & 80 & 2,004,994,693 & 0.98 & 6.05 & 44.84 \\
193 & 500 & Selection + Synthesis & 80 & 2,004,994,693 & 0.87 & 6.48 & 43.77 \\
194 & 1500 & Selection + Synthesis & 80 & 2,004,994,693 & 0.26 & 11.52 & 45.29 \\
195 & 25 & Selection + Synthesis & 90 & 2,255,719,055 & 1.46 & 4.95 & 41.05 \\
196 & 50 & Selection + Synthesis & 90 & 2,255,719,055 & 1.55 & 5.13 & 39.31 \\
197 & 75 & Selection + Synthesis & 90 & 2,255,719,055 & 1.27 & 4.65 & 42.70 \\
198 & 125 & Selection + Synthesis & 90 & 2,255,719,055 & 1.20 & 4.72 & 43.94 \\
199 & 350 & Selection + Synthesis & 90 & 2,255,719,055 & 1.00 & 5.71 & 44.69 \\
200 & 500 & Selection + Synthesis & 90 & 2,255,719,055 & 0.90 & 5.98 & 44.89 \\
201 & 25 & Selection + Synthesis & 100 & 2,507,011,688 & 1.46 & 4.92 & 39.12 \\
202 & 50 & Selection + Synthesis & 100 & 2,507,011,688 & 1.54 & 5.14 & 38.54 \\
203 & 75 & Selection + Synthesis & 100 & 2,507,011,688 & 1.27 & 4.69 & 42.14 \\
204 & 125 & Selection + Synthesis & 100 & 2,507,011,688 & 1.22 & 4.71 & 43.35 \\
205 & 350 & Selection + Synthesis & 100 & 2,507,011,688 & 1.12 & 4.75 & 44.94 \\
206 & 500 & Selection + Synthesis & 100 & 2,507,011,688 & 0.93 & 5.53 & 44.97 \\
207 & 1500 & Selection + Synthesis & 100 & 2,507,011,688 & 0.41 & 9.53 & 45.27 \\
\bottomrule
\end{tabular}
}

\section{Ablation Plots}

\subsection{Performances vs. Data Size and Model Size}

\begin{figure}[h]
\centering    
\includegraphics[width=0.7\columnwidth]{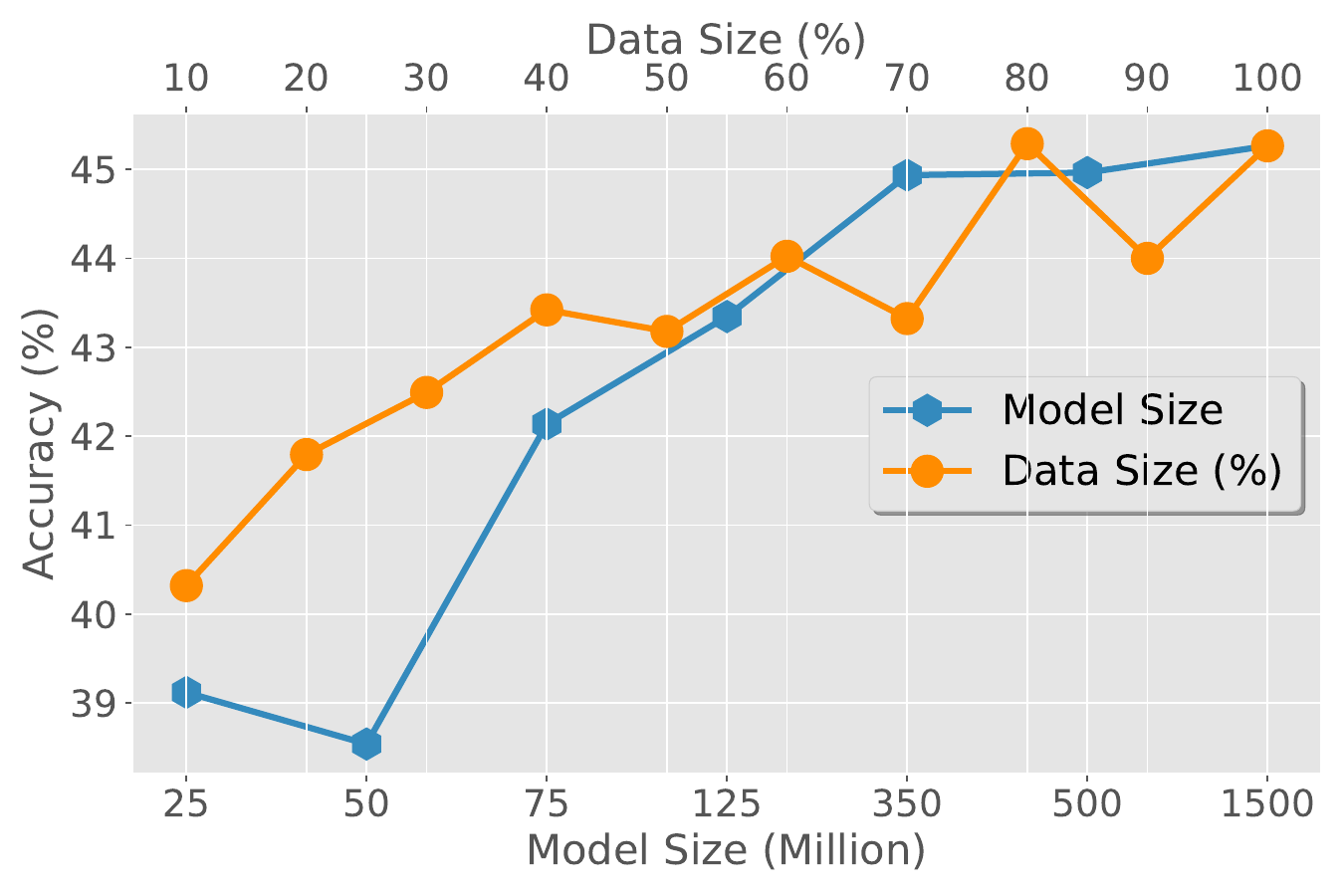}
    \caption{\small Ablating model performances when varying the data sizes (orange) and the model sizes (blue). 
}
\label{fig:ablation}
\end{figure}
\newpage

\subsection{Learning Curve for Varying Model Sizes and Diversity}

\begin{figure}[h]
    \centering
    \includegraphics[width=0.45\columnwidth]{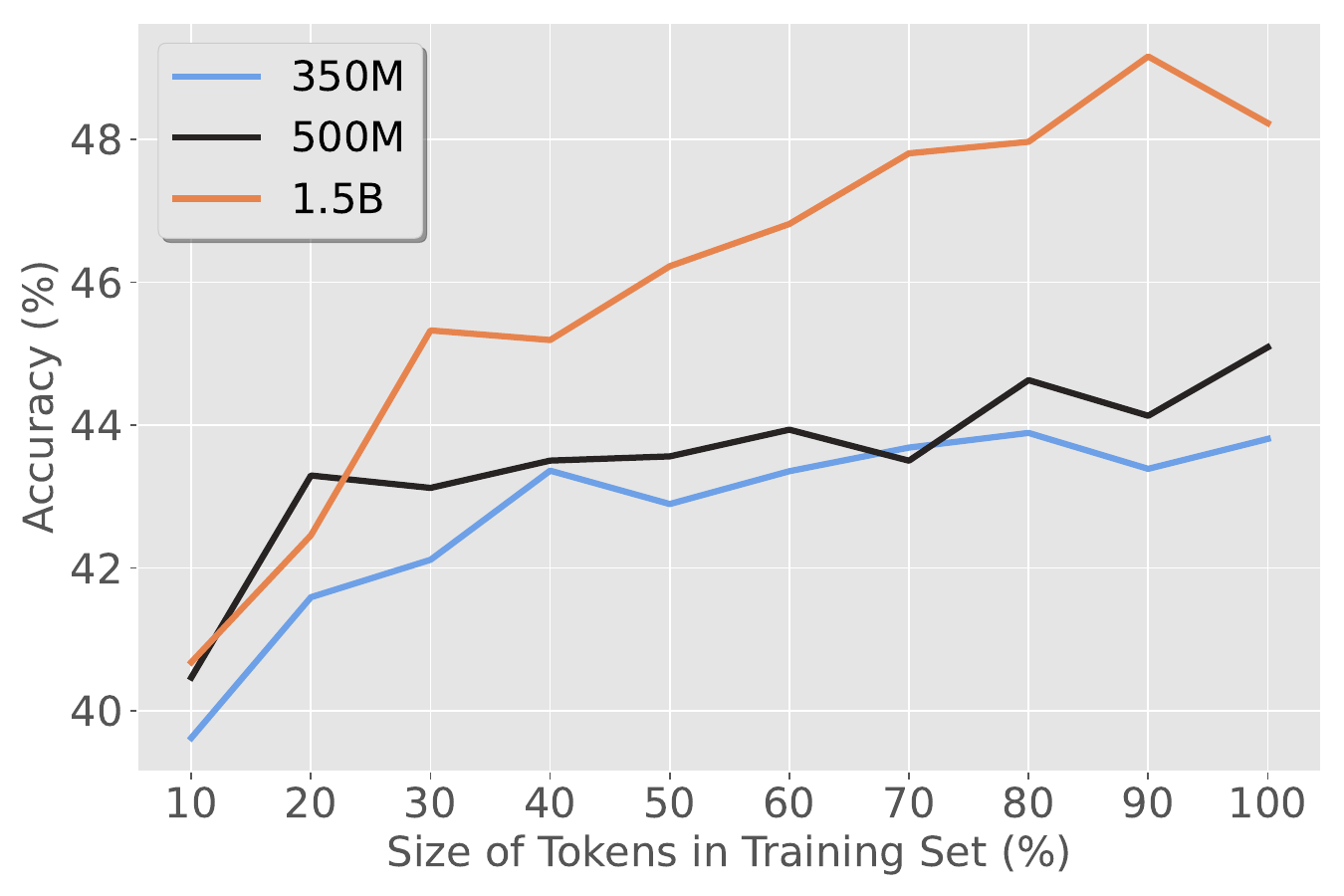}
    
    \caption{\small
    Plot of accuracy against the number of tokens, where tokens are increased in percentages.}   \label{fig:learning_curve_model_size}
\end{figure}

\begin{figure}[th]
  \centering
  \includegraphics[width=0.7\textwidth]{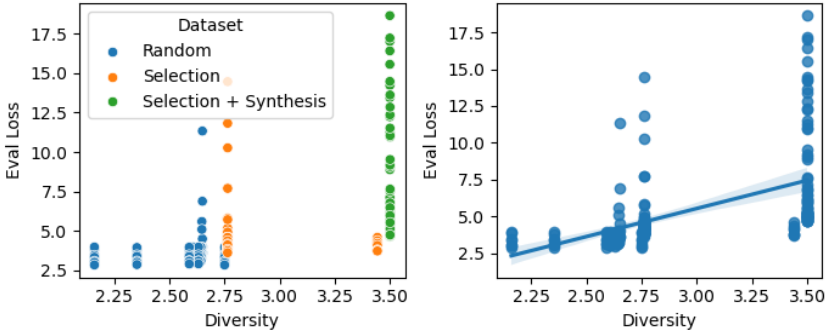}
  \caption{\small Diversity vs. Evaluation Loss: This plot shows the relationship between model diversity and evaluation loss on different datasets.}
  \label{fig:diversity_vs_eval_loss}
\end{figure}
\begin{figure}[th]
  \centering
  \includegraphics[width=0.7\textwidth]{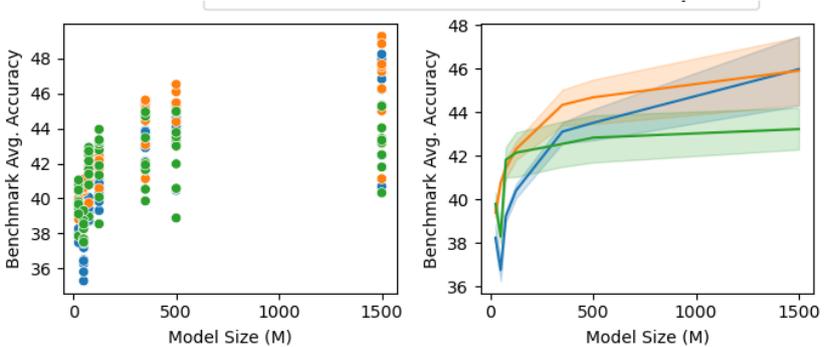}
  \caption{\small  Size vs. Accuracy: This plot shows the relationship between model size and accuracy on different datasets.}
  \label{fig:size_vs_acc}
\end{figure}

\end{document}